\documentclass[runningheads]{llncs}

\usepackage[mobile]{eccv}

\usepackage{eccvabbrv}

\usepackage{graphicx}
\usepackage{booktabs}
\usepackage{times}
\usepackage{microtype}
\usepackage{float}
\usepackage{placeins}
\usepackage{color, colortbl}
\usepackage{stfloats}
\usepackage{enumitem}
\usepackage{tabularx}
\usepackage{xstring}
\usepackage{multirow}
\usepackage{xspace}
\usepackage{url}
\usepackage{lipsum}
\usepackage[titletoc]{appendix}
\usepackage{comment}
\usepackage{bbding}
\usepackage{makecell}
\usepackage{wrapfig}
\usepackage{algorithm}
\usepackage{algpseudocode}
\usepackage{pifont}

\usepackage[accsupp]{axessibility}  

\usepackage{hyperref}

\usepackage[misc]{ifsym}

\definecolor{gold}{rgb}{0.98, 0.83, 0.16}
\definecolor{silver}{rgb}{0.72, 0.72, 0.72}
\newcommand{\best}[1]{\colorbox{gold!50}{#1}}
\newcommand{\secondbest}[1]{\colorbox{silver!50}{#1}}

\definecolor{oneoffour}{RGB}{251,180,174}
\definecolor{twooffour}{RGB}{179,205,227}
\definecolor{threeoffour}{RGB}{204,235,197}
\definecolor{fouroffour}{RGB}{222,203,228}

\newcolumntype{x}{>{\columncolor{oneoffour!80}}c}
\newcolumntype{y}{>{\columncolor{twooffour!80}}c}
\newcolumntype{z}{>{\columncolor{threeoffour!80}}c}
\newcolumntype{w}{>{\columncolor{fouroffour!80}}c}

\makeatletter
\newcommand{\printfnsymbol}[1]{%
        \textsuperscript{\@fnsymbol{#1}}%
}
\makeatother

\begin{document}

\title{RPBG: Towards Robust Neural Point-based Graphics\\ in the Wild} 

\author{Qingtian Zhu\inst{1}\thanks{Work done during Q. Zhu's internship at XREAL.} \and
Zizhuang Wei\inst{2,3} \and
Zhongtian Zheng\inst{3} \and
Yifan Zhan\inst{1} \and 
Zhuyu Yao\inst{4} \and
Jiawang Zhang\inst{4} \and
Kejian Wu\inst{4} \and
Yinqiang Zheng\inst{1}\textsuperscript{\Letter}
}

\authorrunning{Q.~Zhu et al.}
\institute{The University of Tokyo \and
Huawei Technologies \and
Peking University \and
XREAL
}

\maketitle

\begin{abstract}

Point-based representations have recently gained popularity in novel view synthesis, for their unique advantages, \eg, intuitive geometric representation, simple manipulation, and faster convergence.
However, based on our observation, these point-based neural re-rendering methods are only expected to perform well under ideal conditions and suffer from noisy, patchy points and unbounded scenes, which are challenging to handle but 
\textit{defacto} common in real applications.
To this end, we revisit one such influential method, known as Neural Point-based Graphics (NPBG), as our baseline, and propose Robust Point-based Graphics (RPBG).
We in-depth analyze the factors that prevent NPBG from achieving satisfactory renderings on generic datasets, and accordingly reform the pipeline to make it more robust to varying datasets in-the-wild.
Inspired by the practices in image restoration, we greatly enhance the neural renderer to enable the attention-based correction of point visibility and the inpainting of incomplete rasterization, with only acceptable overheads.
We also seek for a simple and lightweight alternative for environment modeling and an iterative method to alleviate the problem of poor geometry.
By thorough evaluation on a wide range of datasets with different shooting conditions and camera trajectories, RPBG stably outperforms the baseline by a large margin, and exhibits its great robustness over state-of-the-art NeRF-based variants.
Code available at \url{https://github.com/QT-Zhu/RPBG}.
\end{abstract}
\section{Introduction}
\begin{figure}[t]
\centering
\includegraphics[width=\linewidth]{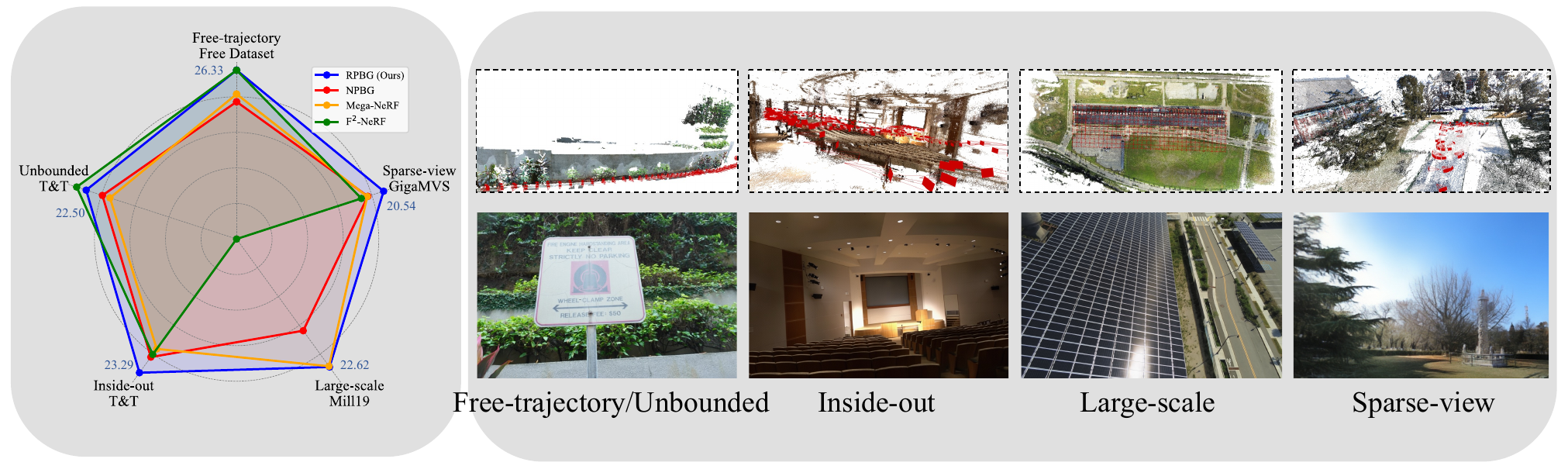}
\caption{\textbf{Left:} RPBG manages to achieve \textbf{all-round} good re-renderings (PSNR plotted) across generic datasets over the baseline~\cite{aliev2020neural} as well as state-of-the-art RF-based methods~\cite{turki2022mega,wang2023f2}.
\textbf{Right:} We demonstrate the point clouds (with camera trajectories visualized) and corresponding re-rendered novel views of the representative scenes, revealing the great robustness and scalability of RPBG. Zoom in for best view.}
\label{fig:teaser}
\end{figure}

Novel view synthesis (NVS) aims to synthesize novel views under given camera poses by aid of a series of already posed images, which is a fundamental task in both computer vision and computer graphics and has been studied for decades.

Among the feasible solutions, NeRF (Neural Radiance Field)~\cite{mildenhall2020nerf} approximates an implicit scene representation, \ie, a radiance field (RF), encoded by a neural network mapping 3D coordinates and view directions to colors and densities.
When rendering, the fitted RF is queried multiple times along a ray to volume-render the corresponding pixel for a novel view.
Subsequent variants of NeRF follow a case-by-case design and adopt different parameterization techniques specifically for different scene types, \eg, object-centric~\cite{zhang2020nerf++,barron2022mip}, free-trajectory~\cite{wang2023f2}, and large-scale scenes~\cite{tancik2022block,turki2022mega}.
Therefore, putting forward one robust method that can work stably across varying scene types with a unified parameterization technique is considered as challenging.

Recently, point-based alternatives~\cite{aliev2020neural,rakhimov2022npbg++,ruckert2022adop,kerbl20233d} have gained substantial attention for their unique advantages over implicit representations, \eg, ease of manipulation and faster training.
The practice of employing point-based parameterization for rendering can be traced all the way back to \cite{levoy1985use,grossman1998point}, and is intuitive especially under the context of NVS.
Analogue to NeRF and its variants (collectively referred to as RF-based methods), whose optimization and convergence heavily rely on the inherent co-visibility among images, the 3D points triangulated from a series of 2D images according to epipolar geometry have already contain all the verified co-visibility information.
In this way, by enforcing the triangulated points as strong prior knowledge for NVS, the expected freedom of optimization is greatly constrained, leading to faster training as well as better robustness.
Recent attempts have achieved promising results in terms of fast optimization and rendering~\cite{kerbl20233d,rakhimov2022npbg++} and accurate geometric representation~\cite{zhang2024papr}.
Concretely, we focus on the specific category of methods that manage to integrate CNN-based neural renderers to yield re-renderings with featuremetric neural buffers~\cite{liu2020neural,rakhimov2022npbg++}, for their concise pipeline and the potential capability of obtaining visually pleasing re-renderings.

However, by evaluating the representative baseline, NPBG (Neural Point-based Graphics)~\cite{aliev2020neural}, with generic datasets, we realize that the original design only intends for ideal conditions, \eg, synthetic data~\cite{mildenhall2020nerf} and well-captured human heads~\cite{ramon2021h3d}.
When synthesizing novel views under more general conditions, the performance of NPBG degrades severely.
In this paper, we strive to boost the robustness of NPBG-like point-based neural re-rendering pipelines and reveal the true potential to achieve state-of-the-art performance across varying datasets in-the-wild, by analyzing the reasons for performance degradation and seeking for remedies.  

Generally speaking, the major difficulties that the vanilla NPBG is faced up with include handling the background, handling patchy point clouds, and identifying correct point visibility.
We reform the CNN-based neural renderer, with an inspiration from image restoration algorithms~\cite{zamir2021multi,cho2021rethinking, chen2022simple} that are able to identify downgraded patterns and restore the corresponding high-quality images.
To make sure the neural renderer can capture sufficient and valid context information with extremely sparse rasterizations, we particularly leverages a Downgrade-aware Convolution (DAC) module to determine the correct point visibility with regard to a given camera pose and performs a pseudo point-wise back-face culling operation with visual self-attention.
The background is also modeled in a lightweight manner.
Instead of incorporating a massive environment map~\cite{ruckert2022adop}, a simple default trainable feature vector can reach similar quantitative results when working with the stronger neural renderer.
We also discover the pseudo density calculated from the neural textures can roughly verify the existence of a given 3D position, which can be used to augment the poorly triangulated point clouds.
In addition, we also simplify the phased training paradigm in \cite{aliev2020neural,rakhimov2022npbg++} and optimize the parameters of both the neural textures and the renderer end-to-end collaboratively.
We term our version of point-based re-rendering as \textbf{RPBG} (Robust Point-based Graphics), to emphasize its great robustness across different generic datasets.

For thorough qualitative and quantitative evaluation, we collect 4 typical challenging scene types as the benchmark for robust NVS, \ie, 360\textdegree~unbounded scenes (with free trajectories), inside-out scenes, large-scale scenes (at the scale of a block or campus), and sparse-view scenes.
As a result, RPBG exhibits great robustness with perceptually satisfactory synthesis across the aforementioned typical scenes types as showcased in \cref{fig:teaser}, where the high-quality re-renderings are obtained with an exactly identical parameterization strategy without any manual configuration, relieving the practitioners from the exhaustive per-scene search of hyper-parameters.
Its stable superiority over state-of-the-art NVS methods, we believe, is of great significance especially for real applications.

The main contributions are three-fold as follows:
\begin{itemize}[noitemsep,topsep=0pt]
  \item We put forward RPBG, as a more robust and practical alternative for re-rendering high-quality images from triangulated 3D points.
  \item According to our in-depth analysis, we enhance the neural re-rendering pipeline regarding the neural renderer, environment modeling, point cloud augmentation, and the training paradigm.
  \item RPBG manages to greatly boost the performance of neural point re-renderer by a large margin, and exhibits stably greater robustness and generalizability over RF-based methods with even better perceptual rendering quality.
\end{itemize}

\section{Related Work}

\paragraph{Radiance Fields for NVS}

Along with the proposed NeRF~\cite{mildenhall2020nerf}, the scene representation of RF is becoming popular for its ease of optimization by the differentiable volume rendering. 
In a RF, each position is assigned with an anisotropic color and a density, and to render a pixel of a novel view, one needs to sample the trained field and conduct ray marching.
The original NeRF~\cite{mildenhall2020nerf}, together with its variants~\cite{zhang2020nerf++,barron2021mip,barron2022mip,tancik2023nerfstudio}, employs an MLP to represent the functional mapping from coordinates and view directions to the radiance values.
Several attempts have been made to encode RFs with different data structures for faster training and inference, \eg, grid voxels~\cite{sun2022direct,fridovich2022plenoxels}, decomposed tensors~\cite{chen2022tensorf}, hash grids~\cite{muller2022instant}.

In practice, the parameterization strategies have a great impact on the rendering quality when reconstructing different types of scenes, \eg, forward-facing ones, 360\textdegree~unbounded ones, and large-scale ones.
For unbounded scenes, NeRF++~\cite{zhang2020nerf++} applies separate networks and different parameterization to model near and distant objects;
mip-NeRF 360~\cite{barron2022mip} designs a smooth contraction operator to parameterize the whole unbounded scene into a ball;
Mega-NeRF~\cite{turki2022mega} and Block-NeRF~\cite{tancik2022block} partition 3D scenes explicitly and use different networks to represent each scene partition;
F$^2$-NeRF~\cite{wang2023f2} proposes to use perspective warping to handle sequential data with arbitrary trajectories.
The varying case-by-case parameterization strategies of RF-based NVS greatly constraint the generalizability of methods.

\paragraph{Point-based Graphics}

A point cloud is a collection of 3D coordinates that is usually used as a topology-agnostic coarse shape representation of the 3D geometry and favored for its flexibility and sparsity for storage.
The development of techniques of employing points as modeling primitives for rendering (referred as point-based graphics~\cite{gross2011point}) can be traced back to \cite{levoy1985use,grossman1998point}, the best practice of which is to replace each point with an oriented circular disk (a surfel) and reply on splatting to blend overlapping surfels~\cite{pfister2000surfels}.

In contrast to conventional physically based rendering (PBR) techniques, neural rendering~\cite{tewari2020state} learns to render high-quality images in a data-driven manner.
Particularly, we focus on the methods that conduct neural re-rendering with point clouds.
Bui \etal~\cite{bui2018point} propose to enhance the coarse point-based rendering by a GAN for image super-resolution.
NRW~\cite{meshry2019neural} and InvSFM~\cite{pittaluga2019revealing} attempt to re-render the reconstructed point cloud with respective auxiliary buffers, \eg, latent appearance vectors, semantic masks and SIFT descriptors~\cite{lowe2004distinctive}.
NPBG~\cite{aliev2020neural,rakhimov2022npbg++} adopts a U-Net-like~\cite{ronneberger2015u} CNN to render neural point textures as RGB images, and exhibits better flexibility over mesh-based proxies~\cite{thies2019deferred}.
The practical applications in the context of autonomous driving, \eg, scene editing and stitching, and large-scale training, are further explored in READ~\cite{li2023read}.
Recently, ADOP~\cite{ruckert2022adop} and Gaussian Splatting~\cite{kerbl20233d} have demonstrated notable accomplishments.
However, the differentiable rendering/splatting scheme employed requires an extensive amount of memory to maintain the computational graph and gradients during training. 
This limitation hinders their application on large-scale scenes.

\section{Neural Point-based Graphics Revisited}

In this section, we revisit the pipeline of NPBG~\cite{aliev2020neural}, analyze the existing drawbacks when attempting to re-render generic scenes, and  attempt to explain why NPBG finds it difficult to handle the background and patchy points, and identify correct point visibility.
We hope these discussions be of sufficient insights to support our modifications.

\subsection{Preliminaries}

\paragraph{Inputs}
As the common practice in NVS, the inputs are a series of images with respective camera parameters (both intrinsics and extrinsics), and specifically for point-based methods~\cite{aliev2020neural,rakhimov2022npbg++,kerbl20233d}, the very first step is supposed to be the triangulation of 3D points from 2D observations.
The original NPBG-like pipelines~\cite{aliev2020neural,rakhimov2022npbg++} mainly consider this as a pre-processing without much attention.

\paragraph{Point Rasterization}
NPBG~\cite{aliev2020neural} and NPBG++~\cite{rakhimov2022npbg++} apply a non-differentiable hard point $z$-buffering operation as an approximated back-face culling when rasterizing points as 2D fragments, where the fragment is updated when and only when the newly projected point has a smaller $z$-depth than the current one.
After a traversal of all the points, the fragment keeps the record of the indices of rasterized points.
Then tensor scattering is performed to index the neural texture at the corresponding positions.
Compared to differentiable rendering/splatting alternatives~\cite{ruckert2022adop,kerbl20233d}, we consider the most significant advantage is its memory efficiency and thus better scalability, since the computation graph required to keep is much smaller for NPBG.
We also want to keep this scalability without interfering the elegant rasterization paradigm.

\paragraph{Neural Renderer}
The renderer $\mathcal{R}_{\Theta}: \mathbb{R}^{H\times W\times C}\rightarrow \mathbb{R}^{H\times W\times 3}$ of NPBG~\cite{aliev2020neural} follows a vanilla U-Net-like architecture~\cite{ronneberger2015u}. 
Since the target point clouds for rendering are supposed to be well-constructed, the vanilla U-Net is expressive enough to complete the assigned task of mapping higher dimensional features from the neural texture $\mathbf{T}$ to RGB values.
Empirically, we find the expected properties of $\mathcal{R}_{\Theta}$ have much in common with a network for low-level vision tasks, \eg, image restoration.

\subsection{Problems}

\begin{figure}[t]
    \centering
    \includegraphics[width=\linewidth]{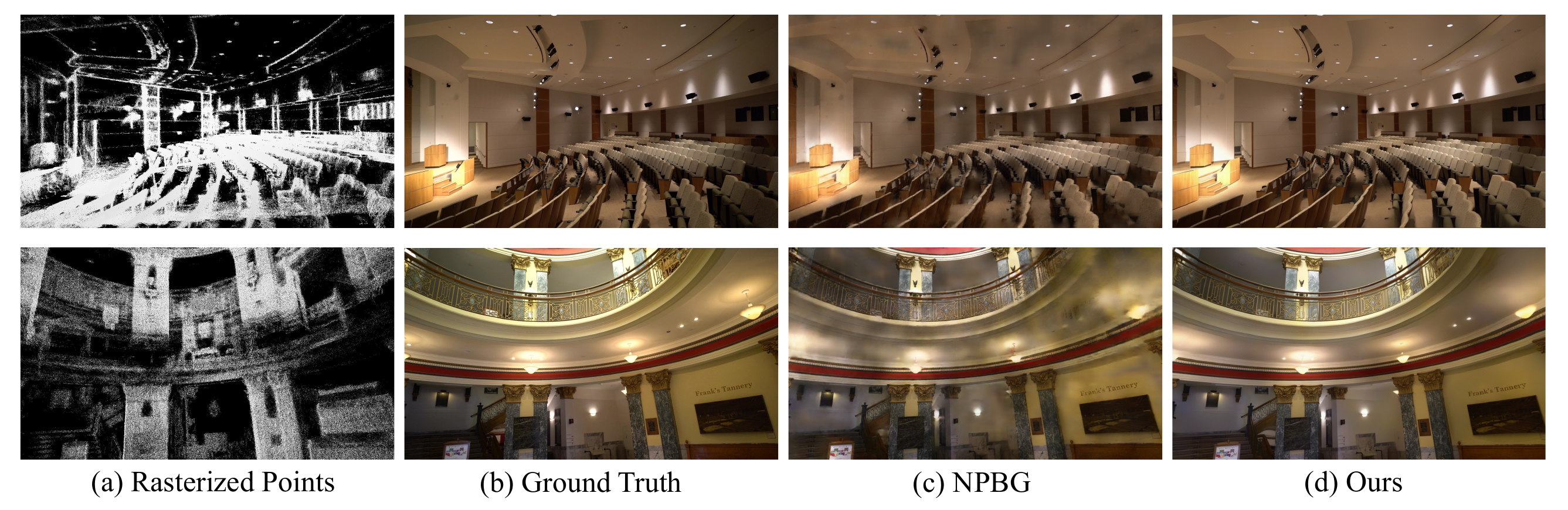}
    \caption{Typically challenging scenes in T\&T dataset~\cite{knapitsch2017tanks} for NPBG. \textbf{Top:} \textit{Auditorium}, where the walls and ceilings are extremely sparse. \textbf{Bottom:} \textit{Museum}, where the point sparsity makes the occlusion and visibility complicated.}
    \label{fig:challenge}
\end{figure}

\paragraph{Patchy Triangulation}
Consider the inputs, for a NVS system aiming to synthesize images for generic scenes, the triangulation of points is not trivial.
The point triangulation step to perform is similar to the step of multi-view stereo (MVS) reconstruction~\cite{furukawa2009accurate,yao2018mvsnet}, the key of which is to identify co-visible pixels across images and lift the 2D pixels to a 3D points according to epipolar geometry.
Such algorithms suffer from non-Lambertian surfaces and textureless regions, both of which are very common and lead to poor triangulated and patchy points.
We attach two typical scenes that NPBG fails to yield good renderings in \cref{fig:challenge}.

\paragraph{Wrong Point Visibility} 
As can be imagined, such point visibility can be erroneous due to the poor quality of triangulated points, and also due to the inherent sparsity of points as a 3D representation.
In this way, the points belonging to back faces, which should be considered as occluded, are rasterized as the fragment for further processing. 
Manually setting a depth threshold for bound check could not be helpful either, since being far from the camera does not necessarily indicate they should be occluded --- they could be parts of the environment.
While NHR~\cite{wu2020multi} proposes to take the depth buffer as an additional rendering condition, we find its effectiveness not significant for generic scenes other than well-masked human captures.

\paragraph{Lack of Context for Re-rendering}
Similarly, when re-rendering an incomplete point cloud, where the neural buffers are usually with significant flaws, the whole receptive field of a kernel at a given position may only capture the downgraded regions, leading to failure to yield reasonable restoration, especially for a high target resolution.
It is worth noting that, though the relationship between the number of network layers and rendering quality has been discussed in \cite{rakhimov2022npbg++,ruckert2022adop}, there is still huge room for the improvement of the renderer.

\paragraph{Failure to Effectively Model the Environment}
Unbounded scenes are a typical challenging scene type for NVS.
RF-based methods that are designed on purpose to handle such scenes~\cite{zhang2020nerf++,barron2022mip} typically employ different parameterization to encode the areas outside a certain bounding sphere.
For point-based methods, ADOP~\cite{ruckert2022adop} proposes to apply an environment map of $H\times W\times C$ to model the environment, which is equivalent to wrapping the triangulated points with a sphere with $H\times W$ points, resulting in more than $5\times 10^5$ points as overheads, according to the default configuration.

\paragraph{Summary}
To summarize, we attribute the observed problems to two main causes: poor geometry and weak, local renderer.
The two causes are to some extent coupled for a better geometry will relieve the difficulty of the renderer and \textit{vice versa}.


\section{Robust Point-based Graphics}
\begin{figure}[t]
    \centering
    \includegraphics[width=\linewidth]{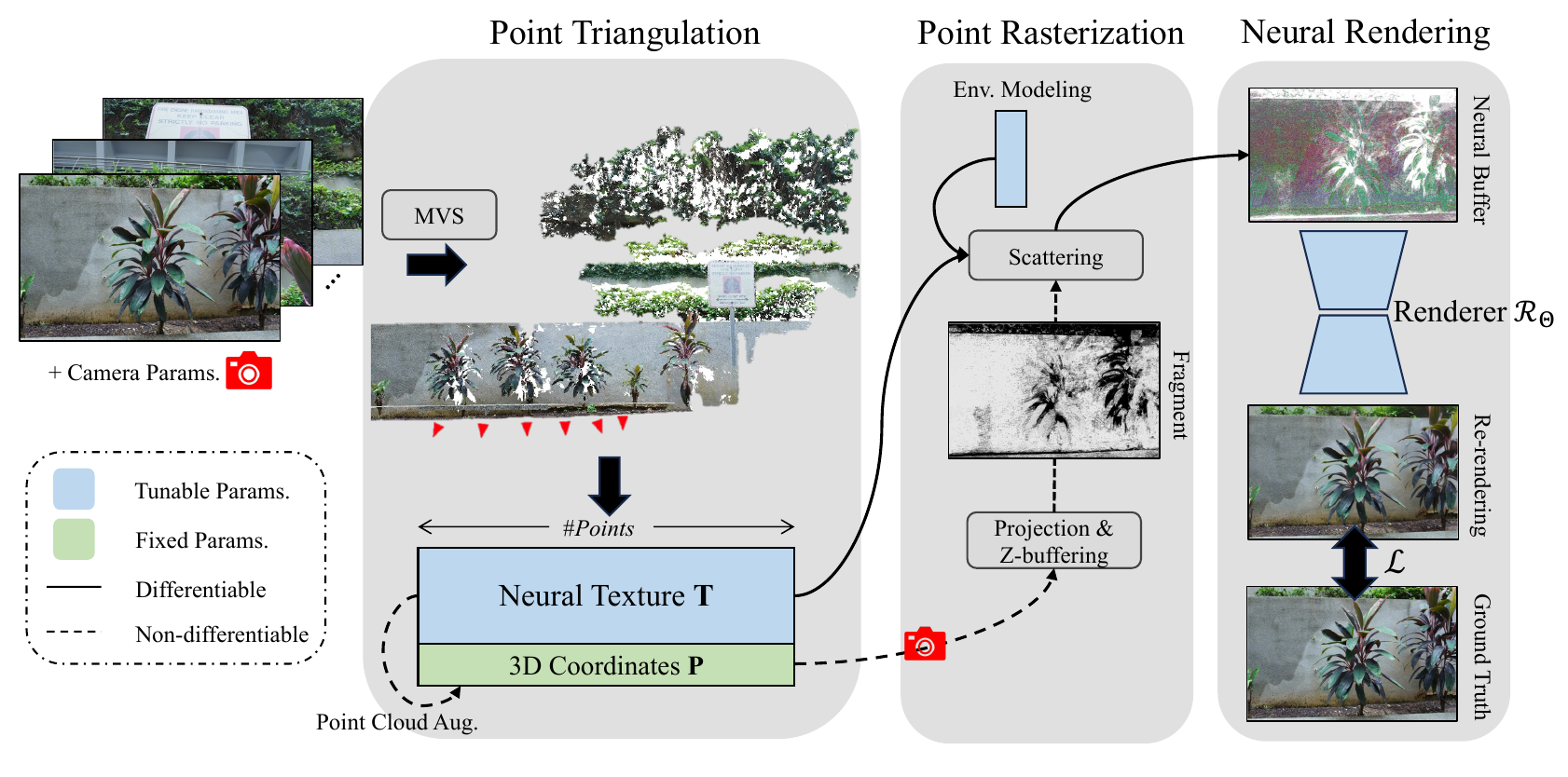}
    \caption{The overall pipeline of RPBG. \textbf{Point Triangulation:} We first triangulate a 3D proxy for re-rendering with posed images, with its geometry-bounded neural texture initialized. \textbf{Point Rasterization:} The points are raterized to the given camera in a non-differentiable manner. By indexing the texture with the fragment, we obtain the neural buffer. A learnable point-size neural texture $\mathbf{T}_{\textrm{env}}$ is also optimized. \textbf{Neural Rendering:} The restoration from downgraded neural buffer to photo-realistic images is performed by a CNN. The network and the neural texture are optimized end-to-end by image-level losses.
    An offline point cloud augmentation strategy is introduced to alleviate the problem of patchy triangulation under challenging conditions.}
    \label{fig:pipeline}
\end{figure}

We are strongly convinced that the pipeline of point-based neural re-rendering has the great potential to outperform popular RF-based solutions, for its unified point-as-parameterization fashion and the incorporation of neural networks to yield visual details.
Therefore in this section, we introduce Robust Point-based Graphics (RPBG), as an enhanced version of NPBG, with a particular focus on robustness under generic scenes.
We will elaborate the modifications made and shed light on the underlying insights and motivation. 
The overall pipeline of RPBG as well as the training paradigm is briefly illustrated in \cref{fig:pipeline}.

\subsection{Downgrade-aware Neural Renderer}
As is attributed as one key problem in NPBG, the U-Net-based neural renderer is considered as too naive to handle the challenging situations in generic scenes.
The modifications made are deeply inspired by low-level vision tasks, where networks can adaptively determine downgraded parts, \eg, the deblurring/deraining networks are able to identify the blurring/raining pixels from the whole image, and correct them accordingly.
We expect the neural renderer $\mathcal{R}_{\Theta}$ can benefit from relevant restoration-targeting techniques, and become able to decode high-quality visual information from patchy buffers.

After evaluating three state-of-the-art fundamental architectures for image restoration, \ie, multi-scale fusion~\cite{cho2021rethinking}, multi-stage~\cite{zamir2021multi}, and U-Net~\cite{chen2022simple}, following the taxonomy in \cite{chen2022simple}.
According to the reported experiments in \cref{tab:ablation_network}, we opt the paradigm in \cite{cho2021rethinking} for it achieves the best balance between performance and time/memory efficiency.
However, the convolution layers in such paradigm is still with a fixed receptive field, lacking in robustness against point sparsity.
We would like to further enlarge the receptive field to a global scale and attempt to explicitly model the visual attention to weigh the observed points.

Transformer-based architectures for image restoration, \eg, Restormer~\cite{zamir2022restormer}, brings unacceptable memory overheads, so we rely on the frequency-domain alternative, Fast Fourier Convolution (FFC)~\cite{chi2020fast}, which can theoretically capture global contexts, to determine the correct point visibility adaptively.
FFC performs channel-wise real 2D FFT (Fast Fourier Transform) and inverse real 2D FFT on 2D tensors.
Real FFT uses only half of the spectrum and by convolving the transformed frequency-domain tensors. 
In this way, a receptive field covering the entire image is considered.

\begin{figure}[t]
\centering
\includegraphics[width=0.8\linewidth]{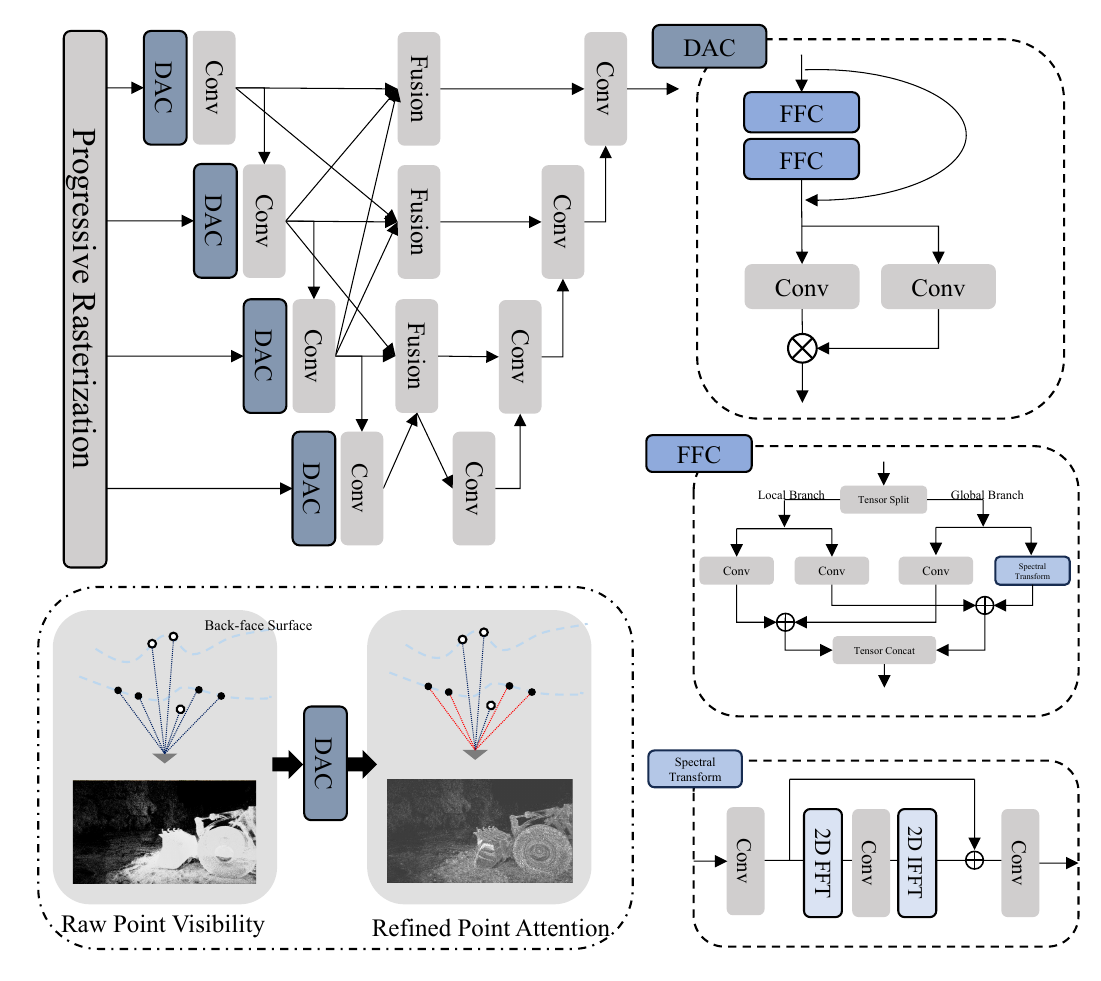}
\caption{The architecture of the downgrade-aware neural renderer in RPBG, with some conventional modules omitted. From the visualized attention map, DAC manages to adaptively handle the severely erroneous point visibility.}
\label{fig:restore}
\end{figure}

Inspired by \cite{szegedy2015going}, we apply FFT (as the global branch) in parallel to conventional convolution layers (as the local branch) and rely on the fused features of both branches to determine the downgraded regions for the gated convolution~\cite{yu2019free} to filter.
Based on the common practice in image inpainting, we leverage gated convolutions at the early stage of the renderer, to help locate the downgrade by wrong point visibility.
We name such customized gated convolution module as the Downgrade-aware Convolution (DAC) module, and as can be inferred from the point attention maps before and after DAC in \cref{fig:restore}, DAC manages to determine complicated point visibility with patchy triangulation.

\subsection{Point Triangulation}
Though there are advances in MVS empowered by deep learning~\cite{wang2021patchmatchnet,wei2021aa,wang2022mvster}, recovering a complete, accurate, and dense point cloud for certain regions remains a challenging. 
For RPBG, we adopt an off-the-shelf MVS method~\cite{wei2021aa} to perform point triangulation for its memory efficiency to allow large-batch inference.

As for problem the poor triangulation, we partially leave it to the neural renderer $\mathcal{R}_{\Theta}$, relying on a stronger renderer to recover the re-rendering from downgraded buffers. 
Inspired by the point cloud extraction method of RF-based implementations~\cite{tancik2023nerfstudio}, where the estimated radiance density $\sigma$ can be a rough indicator for surface, we propose a point cloud augmentation technique to densify the initial triangulation.

The point cloud augmentation follows a trial-and-error paradigm by first assuming the existence of newly sampled points and then verify them by estimated pseudo densities $\sigma_i = \sum |\mathbf{T}_i|$.
It is empirically observed that the absolute activation of the point-wise neural texture can roughly represent the reliability of the 3D position --- it is reasonable since an outlier will be less distinguishable by the neural renderer and thus less visual attention will be given.

Note that this strategy is optional, and we only apply such strategy to scenes with extremely poor triangulation.
Please refer to the Supplementary Material for more discussions regarding the self-pruning.

\subsection{Environment Modeling}

As a result of the incorporation of a stronger neural renderer, we find that given a neural texture with $C=8$, the dense environment map suggested in \cite{ruckert2022adop} is redundant.
Instead, we shrink the overheads for environment modeling from $H\times W\times C$ to $1\times C$, and relying on the stronger neural renderer to decode the background. 

When rasterizing, we employ a tunable feature vector $\mathbf{T}_{\textrm{env}}$ aside the neural point texture $\mathbf{T}$, as the default value for vacant pixels in the fragment, which is also involved in the end-to-end training.
By experiments in \cref{sec:ablation}, we demonstrate that the lightweight modeling strategy reaches the quantitative performance equivalent to applying an environment map, yet with a negligible overhead.

\subsection{Collaborative Optimization}\label{sec:optimization}
Recall that the point rasterization procedure is completely parameter-free. 
The overall collaborative optimization scheme can be thus formulated as
\begin{equation}
    \mathbf{T}^*, \Theta^*  = \mathop{\arg\min}\limits_{\mathbf{T},\Theta} \sum_{k} \mathcal{L}(\mathbf{I}_k, \hat{\mathbf{I}}_k)
     = \mathop{\arg\min}\limits_{\mathbf{T},\Theta} \sum_{k} \mathcal{L}(\mathbf{I}_k,\mathbf{K},\mathbf{R}_k,\mathbf{t}_k|\mathbf{T},\mathcal{R}_{\Theta}),
\end{equation}
where $\mathbf{T}$ and $\Theta$ stand for the tunable parameters, \ie, the neural point texture (alongside the globally shared $\mathbf{T}_{\textrm{env}}$) and the parameters of the renderer $\mathcal{R}_{\Theta}$ for neural re-rendering, while $\mathbf{I}_k$ is the target image whose calibration is $\mathbf{K}$ and $[\mathbf{R}_k|\mathbf{t}_k]$ and $\hat{\mathbf{I}}_k$ is the re-rendering.

Note that, we also discard all the typical tossing sampling~\cite{li2023read} and optimization steps~\cite{rakhimov2022npbg++,ruckert2022adop} of point-based neural re-rendering, and manage to tune all the parameters involved in a simple but effective collaborative end-to-end way.
The neural texture is initialized with all zeros while the rendering CNN is trained from scratch for each scene.

\paragraph{Loss Function}
Since the rendering scheme in both NPBG and RPBG is convolutional, where images are rendered in patches, we are able to apply patch-aware losses to enforce the involvement of neighboring pixels to ensure patch-to-patch consistency, offering perceptually good renderings.
To this end, in addition to the pixel-wise Huber norm $\mathcal{L}_{\textrm{huber}}$ providing the basic supervision and numerical stability, we apply two patch-aware losses to the collaborative optimization of RPBG, namely the perceptual VGG loss~\cite{dosovitskiy2016generating,johnson2016perceptual} $\mathcal{L}_{\textrm{vgg}}$, and the FFT loss~\cite{fuoli2021fourier, cho2021rethinking} $\mathcal{L}_{\textrm{fft}}$.
VGG loss compares the rendered image and the target ground-truth image in a high-dimensional feature space by a pre-trained VGG-19 network~\cite{simonyan2014very}.
It reveals the perceptual similarity between images that pixel-wise metrics fail to measure.
FFT loss measures the image-to-image distance in the frequency domain by carrying out 2D FFT towards images.
The frequency components are considered to be crucial in terms of the perceptual quality~\cite{fuoli2021fourier}.

The final loss function applied in RPBG is composed as 
\begin{equation}
    \mathcal{L} = \lambda_{\textrm{huber}} \mathcal{L}_{\textrm{huber}} + \lambda_{\textrm{vgg}} \mathcal{L}_{\textrm{vgg}} + \lambda_{\textrm{fft}}  \mathcal{L}_{\textrm{fft}}.
\end{equation}

\section{Experiments}
\label{sec:exp}

\subsection{Datasets}

We conduct experiments covering a diverse range of scene types to quantitatively examine the generalizability and robustness of NVS methods, including the Free dataset~\cite{wang2023f2} (7 free-trajectory scans with long camera trajectories), Tanks and Temples (T\&T) dataset~\cite{knapitsch2017tanks} (4 unbounded scans and 5 inside-out scans, without salient dynamic objects), Mill19 dataset~\cite{turki2022mega} (2 large-scale aerial scans with over 1600 images in each scan), and GigaMVS dataset~\cite{zhang2021gigamvs} (8 sparse-view outdoor scans).
Note that for T\&T, we use the raw undistorted images provided by the benchmark without any masking~\cite{liu2020neural}.
We follow the common split protocol, that 1 frame out of every 8 frames is evaluated, for the Free dataset, T\&T dataset, and GigaMVS dataset.
For Mill19 dataset, we apply the officially recommended split to align with the experiments in \cite{turki2022mega}.

We also include some other challenging datasets for a thorough evaluation, \eg, ScanNet++ dataset~\cite{yeshwanth2023scannet++}, a challenging indoor benchmark, OMMO dataset~\cite{lu2023large}, a multi-modal aerial NVS dataset and ETH-MS dataset~\cite{eth_ms_visloc_2021}, a super-large-scale dataset (around 5$k$ images).
Please refer to the Supplementary Material for more quantitative and qualitative results.

\subsection{Implementation Details}

\paragraph{Data Preparation}
To obtain the per-scene point cloud representation for re-rendering, we follow the mapping procedure of COLMAP~\cite{schonberger2016structure} for sparse triangulation and the MVS reconstruction with a trained network~\cite{wei2021aa} for dense triangulation.
Note that the reconstruction only takes a small proportion of time over the whole training phase and we conduct ablation experiments to show that RPBG is robust against different triangulation configurations.
Please refer to \cref{sec:ablation} as well as the Supplementary Material for more information.

\paragraph{Training Settings}

We randomly crop images to square patches of $256\times 256$, with a batch size of 8. The learning rate for the neural point textures is $10^{-1}$, and $10^{-4}$ for the rendering networks, which will decay by a factor of 0.5 if 5 consequent epochs witness no drop in the loss function.
A dimension of 8 is applied for any neural texture regardless of the scene scale or complexity.
For the weights of loss functions, we globally set
$\lambda_{\mathrm{huber}}=10^3$, $\lambda_{\mathrm{vgg}}=1$, and $\lambda_{\mathrm{fft}}=1$.
The training is performed on one NVIDIA GeForce RTX 3090, with a GPU memory consumption of up to 23~GB.
It takes around 8 to 30 GPU hours for training, depending on the data scale.
We would like to emphasize that, RPBG does not require case-by-case scene parameterization or grid search of training hyper-parameters for all the scenes covered in the experiments.

\paragraph{Evaluation Metrics} 
We adopt the metrics of PSNR, SSIM and LPIPS (VGG)~\cite{zhang2018unreasonable} for the evaluation between the synthesized and the target images.
According to \cite{zhang2018unreasonable}, PSNR does not faithfully measure image sharpness and so cannot properly account for the nuances of human visual perception.

\begin{figure*}[t]
\centering
\includegraphics[width=\linewidth]{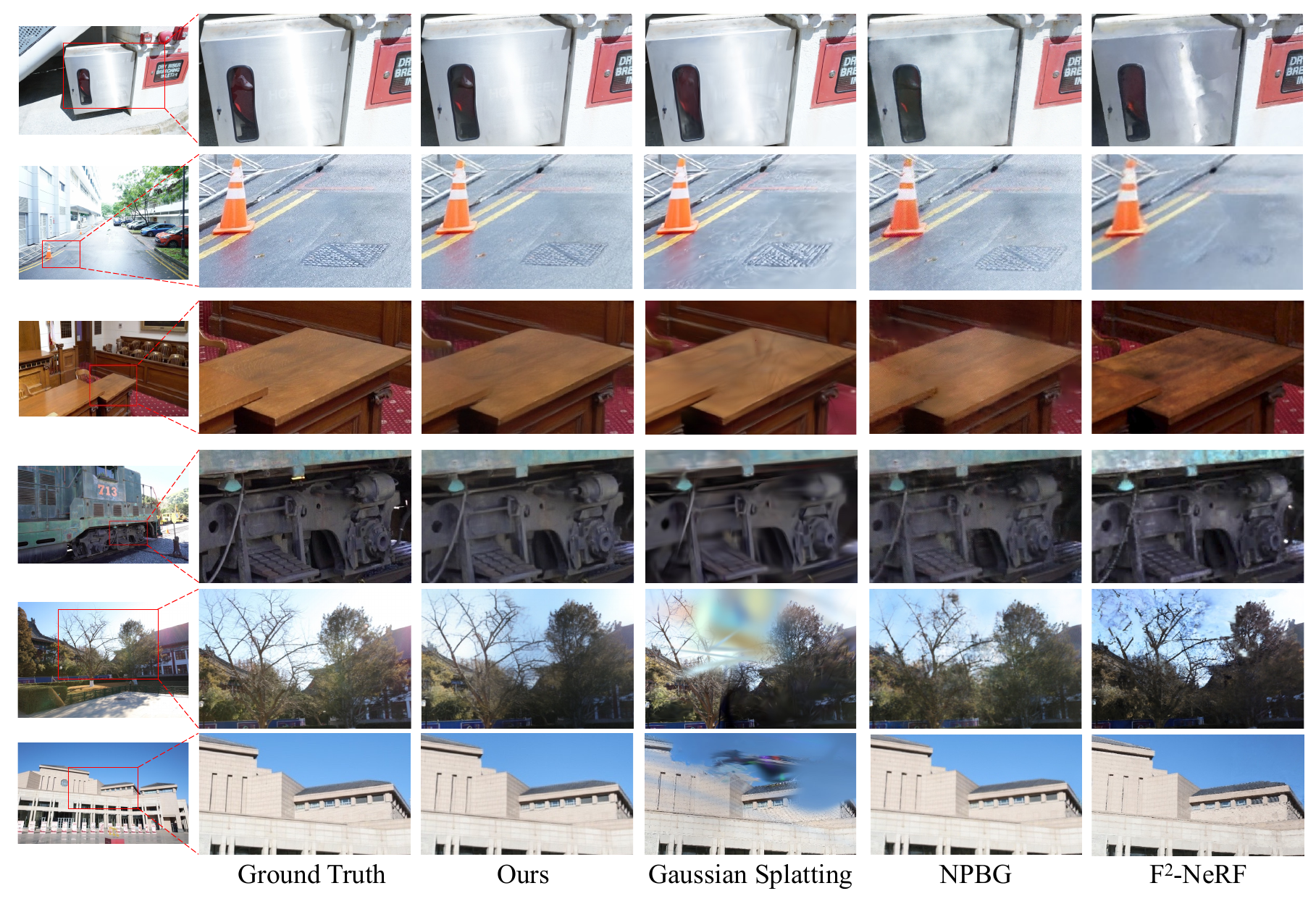}
\caption{Visualized comparisons over varying scenes. \textbf{From top to bottom:} \textit{sky} and \textit{hydrant} of the Free dataset~\cite{wang2023f2}, \textit{Courtroom} and \textit{Train} of T\&T dataset~\cite{knapitsch2017tanks}, and \textit{DayaTemple} and \textit{MemorialHall} of GigaMVS dataset~\cite{zhang2021gigamvs}. We include the results of RPBG (Ours), Gaussian Splatting~\cite{kerbl20233d}, NPBG~\cite{aliev2020neural}, and F$^2$-NeRF~\cite{wang2023f2} for comparison. Zoom in for best view.}
\label{fig:vis-cmp}
\end{figure*}

\begin{table}[t]
    \centering
    \caption{Quantitative evaluation of state-of-the-art NVS methods~\cite{mildenhall2020nerf,chen2022tensorf,zhang2020nerf++,wang2023f2,turki2022mega,aliev2020neural,rakhimov2022npbg++,kerbl20233d} and RPBG across diverse scenes grouped by category, including free trajectory/unbounded scenes, inside-out scenes, large-scale scenes, and sparse-view scenes. The figure following the dataset name stands for the number of scenes the dataset contains. }
    \resizebox{\linewidth}{!}{
    \begin{tabular}{l|ccc|ccc|ccc|ccc|ccc}
        \Xhline{0.8pt}
        \multirow{3}{*}{\textbf{Method}} &   \multicolumn{6}{x|}{\textbf{Free-trajectory/Unbounded}} & \multicolumn{3}{y|}{\textbf{Inside-out}} & \multicolumn{3}{z|}{\textbf{Large-scale}} & \multicolumn{3}{w}{\textbf{Sparse-view}}
        \\
        &  \multicolumn{3}{x|}{Free Dataset-7} & \multicolumn{3}{x|}{T\&T-4} & \multicolumn{3}{y|}{T\&T-5} & \multicolumn{3}{z|}{Mill19-2} & \multicolumn{3}{w}{GigaMVS-8} \\
        &  PSNR$\uparrow$ & SSIM$\uparrow$ & LPIPS$\downarrow$ & PSNR$\uparrow$ & SSIM$\uparrow$ & LPIPS$\downarrow$ & PSNR$\uparrow$ & SSIM$\uparrow$ & LPIPS$\downarrow$ & PSNR$\uparrow$ & SSIM$\uparrow$ & LPIPS$\downarrow$ & PSNR$\uparrow$ & SSIM$\uparrow$ & LPIPS$\downarrow$\\
        \hline
        NeRF~\cite{mildenhall2020nerf}  & 17.75 &0.405 &0.597 & 16.84 & 0.396 & 0.731 & 19.82 & 0.615 & 0.526 & 20.34 & 0.524 & 0.529 & 18.29 &	0.632 &	0.499 \\
        TensoRF~\cite{chen2022tensorf} & 21.74 & 0.549 & 0.600 & 20.73 &	0.643 &	0.566 &	19.43 &	0.634 &	0.570 & 20.07 & 0.497 & 0.597 & 18.70 & 0.653 &	0.506\\
        NeRF++~\cite{zhang2020nerf++} & 23.47 & 0.603 & 0.499  & 21.66 &	0.658 &	0.529 & 19.25 &	0.610 &	0.585 & 20.19 & 0.520 & 0.531 & 18.38 & 0.632 & 0.495 \\
        F$^2$-NeRF~\cite{wang2023f2} & \secondbest{26.32} & \secondbest{0.779} & \secondbest{0.276} & \best{23.66} & 0.764 &	0.303 &	20.12 & 0.706 &	0.394 & N/A & N/A & N/A & 17.44 & 0.540 &	0.470\\
        Mega-NeRF~\cite{turki2022mega}  & 22.60 &	0.570 &	0.562	 & 18.73 &	0.578 &	0.478 & 19.16&	0.617 &	0.451 &  \secondbest{22.49} & \secondbest{0.550} & \secondbest{0.510} & 18.25 &	0.581 &	0.394 \\
        \hline
        NPBG~\cite{aliev2020neural} & 21.40 & 0.639 & 0.340 & 19.85 &	0.698 &	0.376 &	20.57 &	0.696 &	0.371 & 16.21 & 0.357 & 0.644 & 18.34 &	0.620 &	0.405\\
        NPBG++~\cite{rakhimov2022npbg++}  & 20.06 & 0.592 & 0.445 & 17.23 &	0.653 &	0.474 &	18.30 &	0.684 &	0.411 & 17.04 & 0.400 & 0.648 & \secondbest{19.30} & \secondbest{0.663} & 0.443 \\
         Gaussian Splatting~\cite{kerbl20233d} & 25.23 &	0.740 &	0.290 & \secondbest{23.51} &	\best{0.782} &	\secondbest{0.293} &	\best{23.46} &	\secondbest{0.783} &	\secondbest{0.277} & N/A & N/A & N/A & 16.84 &	0.530 & \secondbest{0.391}\\
        \hline
        \textbf{Ours} &  \best{26.33} & \best{0.832} & \best{0.177} & 22.50 &	\best{0.782} &	\best{0.276} &	\secondbest{23.29} &	\best{0.804} &	\best{0.242} & \best{22.62} & \best{0.596} & \best{0.368} & \best{20.54} & \best{0.686} & \best{0.317} \\
        \Xhline{0.8pt}
    \end{tabular}}
    \label{tab:category_cmp}
\end{table}

\subsection{Results}

In \cref{fig:vis-cmp}, we demonstrate several typical groups of visual comparisons of NVS results  by RPBG, two representative point-based methods, \ie, Gaussian Splatting~\cite{kerbl20233d} and NPBG~\cite{aliev2020neural}, and a state-of-the-art NeRF variant for unbounded scenes, F$^2$-NeRF~\cite{wang2023f2}.
We group the datasets by category and report the corresponding scene-averaged quantitative scores in \cref{tab:category_cmp}.
Note that RPBG achieves the best SSIM and LPIPS, which are more relevant to the high-frequency components of images, across all datasets.

\paragraph{Free-trajectory/Unbounded Scenes}
Compared with NVS methods designed in particular for unbounded scenes, NeRF++~\cite{zhang2020nerf++} and F$^2$-NeRF~\cite{wang2023f2}, RPBG achieves the best SSIM and LPIPS and comparable PSNR.
For visual effects showcased in \cref{fig:vis-cmp}, since RPBG involves a neighborhood for rendering, its results appear more visually harmonious, especially than Gaussian Splatting~\cite{kerbl20233d} and F$^2$-NeRF~\cite{wang2023f2}.
It is necessary to mention that, for RF-based methods~\cite{mildenhall2020nerf,chen2022tensorf,turki2022mega,wang2023f2}, we have manually adjusted the scene-specific hyper-parameters to achieve better results.

\paragraph{Inside-out Scenes}
The inside-out indoor scenes are strictly bounded, but  lack high-quality ray intersections required for optimization, which will lead to under-fitting of the RF (\textit{Courtroom} in \cref{fig:vis-cmp}).
Similar to free-trajectory/unbounded scenes, RPBG outperforms state-of-the-art methods in SSIM and LPIPS.

\paragraph{Large-scale Scenes}

We mainly evaluate RPBG against Mega-NeRF~\cite{turki2022mega} on the massive imagery of Mill19 dataset~\cite{turki2022mega}.
Our method outperforms Mega-NeRF~\cite{turki2022mega} in every metric.
It is worth noting that Mega-NeRF takes 240 GPU hours for training while RPBG only takes 29 GPU hours. 
Notably, our system with 256~GB RAM and RTX 3090 fails to afford the training of F$^2$-NeRF~\cite{wang2023f2} and Gaussian Splatting~\cite{kerbl20233d} (marked as N/A in the table). 
In contrast, RPBG proves to be viable for large-scale scenes under the same hardware constraint.

\paragraph{Sparse-view Scenes}
Sparse-view inputs are considered to be extremely challenging for NVS methods.
Since RPBG, NPBG~\cite{aliev2020neural}, and NPBG++~\cite{rakhimov2022npbg++} incorporate a point-based 3D proxy for re-rendering, they showcase better robustness over RF-based methods.
As revealed in \cref{fig:vis-cmp}, Gaussian Splatting~\cite{kerbl20233d} fails to regularize the strong approximation power of point-wise Gaussians at training views, and yields obvious needle-like artifacts at novel views.

\subsection{Ablation Study}\label{sec:ablation}

\begin{wrapfigure}{r}{0.6\textwidth}
  \begin{center}
  \vspace{-40pt}
    \includegraphics[width=0.6\textwidth]{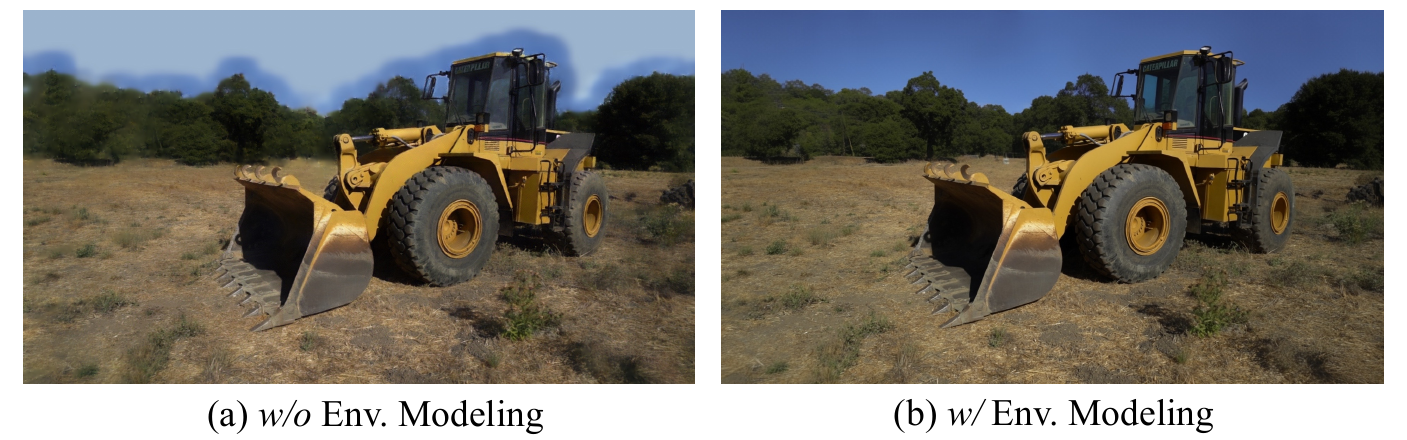}
  \end{center}
  \vspace{-15pt}
  \caption{The impact of leveraging the default feature vector for environment modeling in an unbounded scene.}
  \label{fig:env}
  \vspace{-30pt}
\end{wrapfigure}
\paragraph{Environment Modeling}
We study different strategies of environment modeling for NVS in-the-wild, namely leaving the blank pixels with zeros, filling the blanks with a default learnable feature (as is done in RPBG), and wrapping the points with a sphere of $10^6$ points (equivalent to an environment map in \cite{ruckert2022adop}).
The quantitative results are in \cref{tab:ablation_env}, where the strategy incorporated by RPBG boosts the performance by a large margin, with almost no additional overhead.

\begin{figure}[t]
\begin{minipage}[t]{0.49\linewidth}
\centering
\vspace{0pt}
\includegraphics[width=0.99\linewidth]{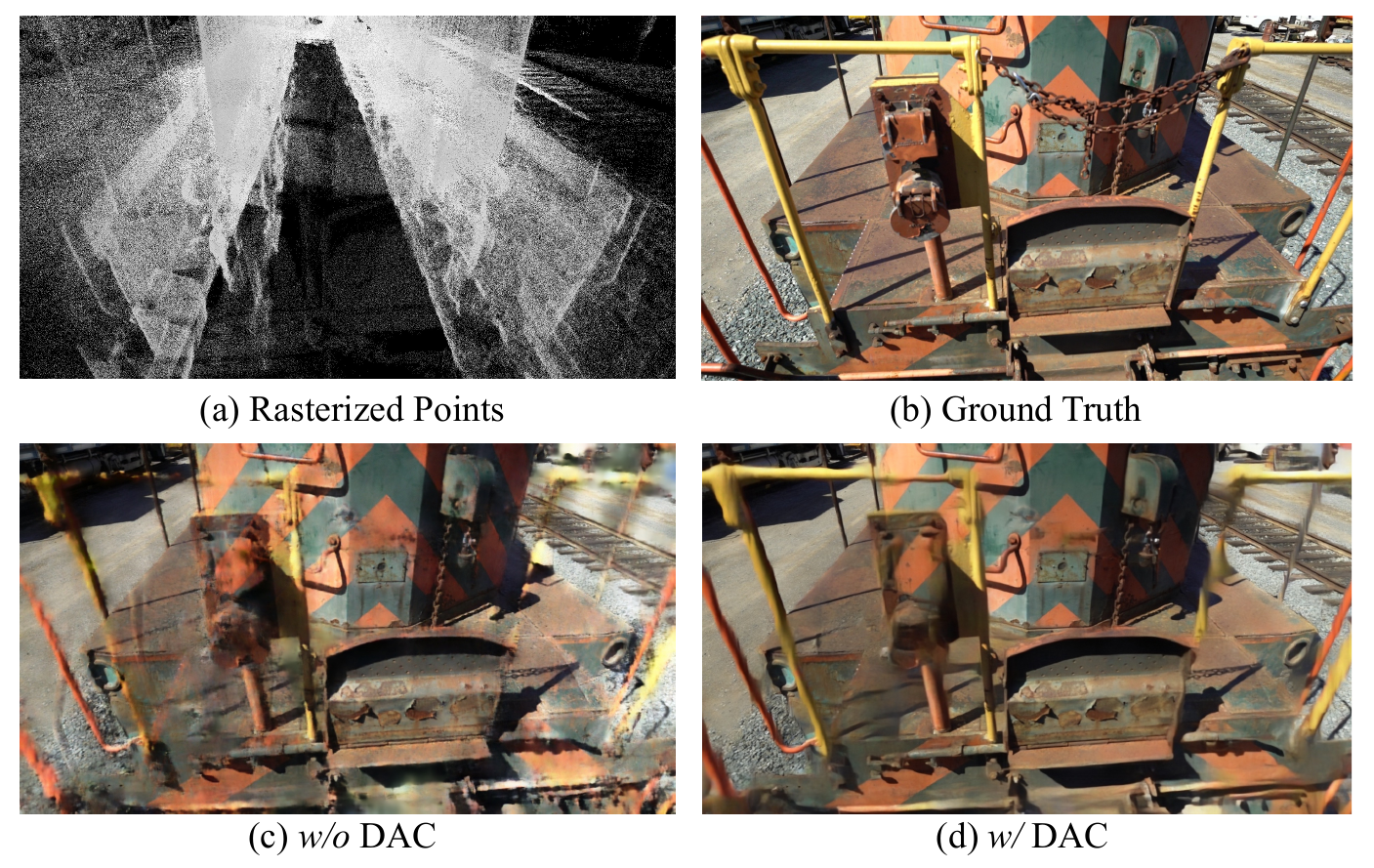}
\caption{A challenging case in \textit{Train}, which is a zoomed-in view with complex occlusion and thin objects (\eg, the handrails).}
\label{fig:abltion_dac}
\end{minipage}
\hfill
\begin{minipage}[t]{0.49\linewidth}
\centering
\vspace{0pt}
\includegraphics[width=0.99\linewidth]{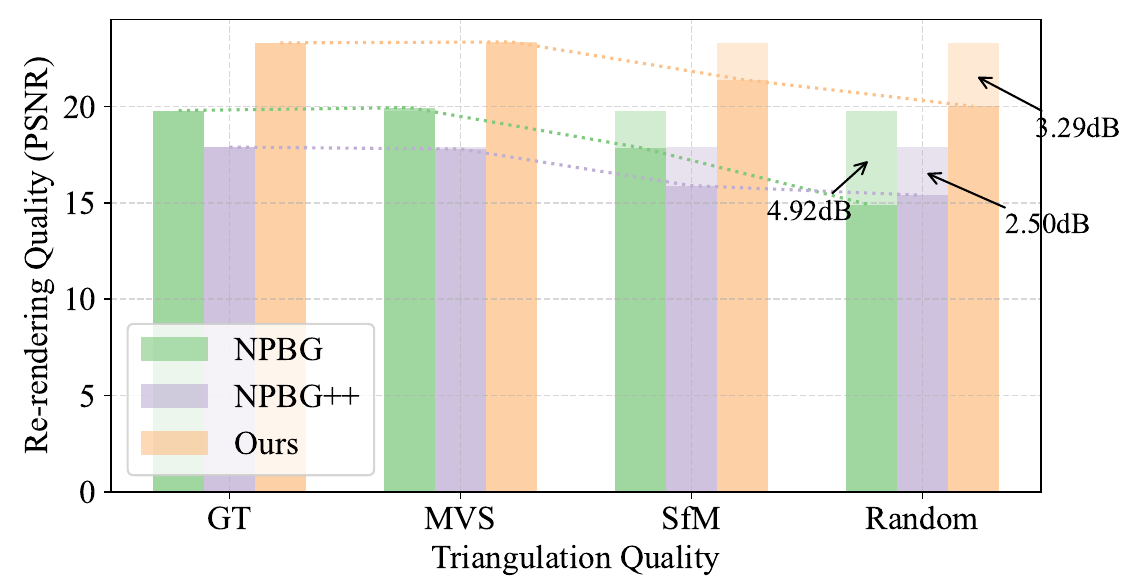}
\caption{Re-rendering quality (PSNR) with points of different quality on \textit{Church}. RPBG manages to maintain its robustness against different levels of point sparsity with a 14\% drop in PSNR (NPBG~\cite{aliev2020neural}: 25\%; NPBG++~\cite{rakhimov2022npbg++}: 14\%).}
\label{fig:bar}
\end{minipage}
\end{figure}

\paragraph{Network Architecture and Loss}
We study the key components in the rendering network in \cref{tab:ablation_network}.
The baseline refers to NPBG~\cite{aliev2020neural} trained per-scene from scratch with $\mathcal{L}_1$ and VGG loss.
We compare three fundamental architectures for image restoration (with necessary modifications on the first several layers), \ie, multi-scale fusion~\cite{cho2021rethinking}, multi-stage~\cite{zamir2021multi}, and U-Net~\cite{chen2022simple} following the taxonomy in \cite{chen2022simple}.
Compared to the baseline, all modern restoration-oriented architectures bring remarkable improvement to the overall performance, and \cite{cho2021rethinking} is opted for a better balance between performance and time/memory efficiency.
The gain by DAC is also considerable while the FFT loss mainly improves the perceptual quality.
By the case in \cref{fig:abltion_dac}, we show the necessity of the DAC module when re-rendering challenging cases without in particular handling the erroneous point visibility.

\begin{table}[t]
\begin{minipage}[t]{0.49\linewidth}
\centering
\label{tab:ablation_network}
\caption{Ablation on network designs on T\&T dataset~\cite{knapitsch2017tanks}. 
    The reported inference time and memory consumption is tested with a target resolution of $1920\times 1080$.}
    \resizebox{0.99\linewidth}{!}{
    \begin{tabular}{l|cc|ccc}
        \Xhline{0.8pt}
        \textbf{Method} & Time($s$) & Mem.(GB)  & PSNR$\uparrow$ & SSIM$\uparrow$ & LPIPS$\downarrow$\\
        \hline
        Baseline & 1.03 & 5.49 & 20.01 & 0.666 & 0.390\\
        +Multi-scale fusion & 1.62 & 12.67 & 21.88 & 0.701 & 0.337\\
        ~~+DAC & 1.67 & 13.01 & 22.92 & 0.769 & 0.320  \\
        ~~~~+FFT loss & -- & -- & 22.94 & 0.794 & 0.257\\
        +Multi-stage & 1.77 & 17.14 & 21.75 & 0.713 & 0.338 \\
        +U-Net & 1.58 & 10.32 & 20.85 & 0.692 & 0.364\\
        \Xhline{0.8pt}
    \end{tabular}}
\end{minipage}\hfill
\begin{minipage}[t]{0.49\linewidth}
\centering
\label{tab:ablation_env}
    \caption{Ablation on environment modeling on two unbounded scenes, namely \textit{Caterpillar} of T\&T dataset~\cite{knapitsch2017tanks} and \textit{sky} of Free dataset~\cite{wang2023f2}. \#\textit{Points} ($\times 10^6$) represent the equivalent overheads when applying different strategies.}
    \resizebox{0.99\linewidth}{!}{
    \begin{tabular}{c|cccc|cccc}
        \Xhline{0.8pt}
        \multirow{2}{*}{\textbf{Env.}} & \multicolumn{4}{c|}{\textit{\textbf{Caterpillar}}} & \multicolumn{4}{c}{\textit{\textbf{sky}}}\\
        & \#\textit{Points} &  PSNR$\uparrow$ & SSIM$\uparrow$ & LPIPS$\downarrow$ & \#\textit{Points} & PSNR$\uparrow$ & SSIM$\uparrow$ & LPIPS$\downarrow$\\
        \hline
        Zeros & 7.95 & 19.89 & 0.456 & 0.403 & 17.49 & 21.77 & 0.598 & 0.315\\ 
         Learnable & 7.95 & 21.78 & 0.687 & 0.289 & 17.49 & 24.81 & 0.866 & 0.199\\
         Sphere & 8.95 & 21.68 & 0.686 & 0.283 & 18.49 & 25.15 & 0.869 & 0.179\\
        \Xhline{0.8pt}
    \end{tabular}}
\end{minipage}
\end{table}

\paragraph{Triangulation Configurations}
We also evaluate RPBG with different triangulation configurations.
Indicated by the results in \cref{fig:bar}, RPBG witnesses a total drop of 3.29~dB in re-rendering PSNR when switching the triangulation from ground truth (GT) to randomly initialized points (Random).
Note that even with sparse points (SfM), which are usually yielded as a by-product when computing camera parameters, RPBG (21.41~dB) still outperforms F$^2$-NeRF (20.57~dB).

\section{Discussion}

\paragraph{Effectiveness of RPBG}
We would like to shed some light on the effectiveness of RPBG.
The first factor should be the point representation.
The triangulated 3D points have already contained all the verified co-visibility that is required for NVS.
Compared to RF-based methods, which often struggle at local minima, such triangulation-as-parameterization paradigm greatly lower the complexity of searching, and is unified across different scene types.
Besides, the patch-wise rendering scheme mentioned in \cref{sec:optimization} also improves the perceptual quality of re-rendering while RF-based methods apply pixel/ray-wise volume rendering, where the correlation between pixels are not modeled explicitly.
For more discussions on the insights, we strongly refer the readers to the Supplementary Material.

\paragraph{Limitations}

An obvious problem with point-based re-rendering (RPBG and NPBG~\cite{aliev2020neural}) is that compared to lightweight RF-based variants~\cite{chen2022tensorf,muller2022instant}, they take more space to store the CNN parameters and the neural texture.
Although it makes the scaling-up easier, the number of parameters grows linearly with the point cloud scale. 
Besides, due to the non-local neural renderer we employ, each point is encoded with visual context of a larger range, hindering the editability of RPBG.
We have also noticed that, the CNN-based rendering scheme can lead to unsatisfactory temporal consistency especially when the triangulation is patchy so that we mainly count on the inpainting ability of the neural render for yielding re-renderings.
Such flicker issue, as well as its potential solutions, has been discussed in \cite{dai2020neural}.
\section{Conclusion}

In this paper, we present RPBG as a robust and practical alternative to NPBG, a baseline of point-based NVS methods, performing neural re-rendering on triangulated points.
We analyze the key problems in NPBG, when attempting to generalize to more generic scenes other than the well-captured scans, and reform the pipeline to reveal the real potential of point-based graphics.
Respectively motivated and inspired by RF-based methods and low-level image restoration methods, we reform the pipeline according to our analysis. 
By extensive experiments on diverse datasets, RPBG achieves stably superior results over state-of-the-art RF-based and point-based NVS methods, especially on the metrics with more attention paid to the nuances of human visual perception, without case-by-case parameterization across all scenes, indicating its robustness and generalizability.

\newpage
\section*{Acknowledgments}
This research was supported in part by JSPS KAKENHI Grant Numbers 24K22318, 22H00529, 20H05951, and JST-Mirai Program JPMJMI23G1.

\bibliographystyle{splncs04}
\bibliography{egbib}

\begin{thebibliography}{10}
\providecommand{\url}[1]{\texttt{#1}}
\providecommand{\urlprefix}{URL }
\providecommand{\doi}[1]{https://doi.org/#1}

\bibitem{aliev2020neural}
Aliev, K.A., Sevastopolsky, A., Kolos, M., Ulyanov, D., Lempitsky, V.: Neural point-based graphics. In: Computer Vision--ECCV 2020: 16th European Conference, Glasgow, UK, August 23--28, 2020, Proceedings, Part XXII 16. pp. 696--712. Springer (2020)

\bibitem{barnes2009patchmatch}
Barnes, C., Shechtman, E., Finkelstein, A., Goldman, D.B.: Patchmatch: A randomized correspondence algorithm for structural image editing. ACM Trans. Graph.  \textbf{28}(3), ~24 (2009)

\bibitem{barron2021mip}
Barron, J.T., Mildenhall, B., Tancik, M., Hedman, P., Martin-Brualla, R., Srinivasan, P.P.: Mip-nerf: A multiscale representation for anti-aliasing neural radiance fields. In: Proceedings of the IEEE/CVF International Conference on Computer Vision. pp. 5855--5864 (2021)

\bibitem{barron2022mip}
Barron, J.T., Mildenhall, B., Verbin, D., Srinivasan, P.P., Hedman, P.: Mip-nerf 360: Unbounded anti-aliased neural radiance fields. In: Proceedings of the IEEE/CVF Conference on Computer Vision and Pattern Recognition. pp. 5470--5479 (2022)

\bibitem{bui2018point}
Bui, G., Le, T., Morago, B., Duan, Y.: Point-based rendering enhancement via deep learning. The Visual Computer  \textbf{34},  829--841 (2018)

\bibitem{openmvs2020}
Cernea, D.: {OpenMVS}: Multi-view stereo reconstruction library (2020), \url{https://cdcseacave.github.io/openMVS}

\bibitem{chen2022tensorf}
Chen, A., Xu, Z., Geiger, A., Yu, J., Su, H.: Tensorf: Tensorial radiance fields. In: European Conference on Computer Vision. pp. 333--350. Springer (2022)

\bibitem{chen2021mvsnerf}
Chen, A., Xu, Z., Zhao, F., Zhang, X., Xiang, F., Yu, J., Su, H.: Mvsnerf: Fast generalizable radiance field reconstruction from multi-view stereo. In: Proceedings of the IEEE/CVF International Conference on Computer Vision. pp. 14124--14133 (2021)

\bibitem{chen2022simple}
Chen, L., Chu, X., Zhang, X., Sun, J.: Simple baselines for image restoration. In: European Conference on Computer Vision. pp. 17--33. Springer (2022)

\bibitem{chi2020fast}
Chi, L., Jiang, B., Mu, Y.: Fast fourier convolution. Advances in Neural Information Processing Systems  \textbf{33},  4479--4488 (2020)

\bibitem{cho2021rethinking}
Cho, S.J., Ji, S.W., Hong, J.P., Jung, S.W., Ko, S.J.: Rethinking coarse-to-fine approach in single image deblurring. In: Proceedings of the IEEE/CVF international conference on computer vision. pp. 4641--4650 (2021)

\bibitem{dai2020neural}
Dai, P., Zhang, Y., Li, Z., Liu, S., Zeng, B.: Neural point cloud rendering via multi-plane projection. In: Proceedings of the IEEE/CVF Conference on Computer Vision and Pattern Recognition. pp. 7830--7839 (2020)

\bibitem{deng2022depth}
Deng, K., Liu, A., Zhu, J.Y., Ramanan, D.: Depth-supervised nerf: Fewer views and faster training for free. In: Proceedings of the IEEE/CVF Conference on Computer Vision and Pattern Recognition. pp. 12882--12891 (2022)

\bibitem{dosovitskiy2016generating}
Dosovitskiy, A., Brox, T.: Generating images with perceptual similarity metrics based on deep networks. Advances in neural information processing systems  \textbf{29} (2016)

\bibitem{eth_ms_visloc_2021}
{ETH Zurich Computer Vision Group and Microsoft Mixed Reality \& AI Lab Zurich}: {The ETH-Microsoft Localization Dataset}. \url{https://github.com/cvg/visloc-iccv2021} (2021)

\bibitem{fridovich2022plenoxels}
Fridovich-Keil, S., Yu, A., Tancik, M., Chen, Q., Recht, B., Kanazawa, A.: Plenoxels: Radiance fields without neural networks. In: Proceedings of the IEEE/CVF Conference on Computer Vision and Pattern Recognition. pp. 5501--5510 (2022)

\bibitem{fuoli2021fourier}
Fuoli, D., Van~Gool, L., Timofte, R.: Fourier space losses for efficient perceptual image super-resolution. In: Proceedings of the IEEE/CVF International Conference on Computer Vision. pp. 2360--2369 (2021)

\bibitem{furukawa2009accurate}
Furukawa, Y., Ponce, J.: Accurate, dense, and robust multiview stereopsis. IEEE transactions on pattern analysis and machine intelligence  \textbf{32}(8),  1362--1376 (2009)

\bibitem{gross2011point}
Gross, M., Pfister, H.: Point-based graphics. Elsevier (2011)

\bibitem{grossman1998point}
Grossman, J.P., Dally, W.J.: Point sample rendering. In: Rendering Techniques’ 98: Proceedings of the Eurographics Workshop in Vienna, Austria, June 29—July 1, 1998 9. pp. 181--192. Springer (1998)

\bibitem{jancosek2014exploiting}
Jancosek, M., Pajdla, T.: Exploiting visibility information in surface reconstruction to preserve weakly supported surfaces. International scholarly research notices  \textbf{2014} (2014)

\bibitem{johnson2016perceptual}
Johnson, J., Alahi, A., Fei-Fei, L.: Perceptual losses for real-time style transfer and super-resolution. In: Computer Vision--ECCV 2016: 14th European Conference, Amsterdam, The Netherlands, October 11-14, 2016, Proceedings, Part II 14. pp. 694--711. Springer (2016)

\bibitem{kerbl20233d}
Kerbl, B., Kopanas, G., Leimk{\"u}hler, T., Drettakis, G.: 3d gaussian splatting for real-time radiance field rendering. ACM Transactions on Graphics (TOG)  \textbf{42}(4),  1--14 (2023)

\bibitem{knapitsch2017tanks}
Knapitsch, A., Park, J., Zhou, Q.Y., Koltun, V.: Tanks and temples: Benchmarking large-scale scene reconstruction. ACM Transactions on Graphics (ToG)  \textbf{36}(4),  1--13 (2017)

\bibitem{levoy1985use}
Levoy, M., Whitted, T.: The use of points as a display primitive  (1985)

\bibitem{li2023read}
Li, Z., Li, L., Zhu, J.: Read: Large-scale neural scene rendering for autonomous driving. In: Proceedings of the AAAI Conference on Artificial Intelligence. vol.~37, pp. 1522--1529 (2023)

\bibitem{liu2020neural}
Liu, L., Gu, J., Zaw~Lin, K., Chua, T.S., Theobalt, C.: Neural sparse voxel fields. Advances in Neural Information Processing Systems  \textbf{33},  15651--15663 (2020)

\bibitem{lowe2004distinctive}
Lowe, D.G.: Distinctive image features from scale-invariant keypoints. International journal of computer vision  \textbf{60},  91--110 (2004)

\bibitem{lu2023large}
Lu, C., Yin, F., Chen, X., Chen, T., Yu, G., Fan, J.: A large-scale outdoor multi-modal dataset and benchmark for novel view synthesis and implicit scene reconstruction. arXiv preprint arXiv:2301.06782  (2023)

\bibitem{meshry2019neural}
Meshry, M., Goldman, D.B., Khamis, S., Hoppe, H., Pandey, R., Snavely, N., Martin-Brualla, R.: Neural rerendering in the wild. In: Proceedings of the IEEE/CVF Conference on Computer Vision and Pattern Recognition. pp. 6878--6887 (2019)

\bibitem{mildenhall2020nerf}
Mildenhall, B., Srinivasan, P.P., Tancik, M., Barron, J.T., Ramamoorthi, R., Ng, R.: Nerf: Representing scenes as neural radiance fields for view synthesis. In: European Conference on Computer Vision. pp. 405--421 (2020)

\bibitem{muller2022instant}
M{\"u}ller, T., Evans, A., Schied, C., Keller, A.: Instant neural graphics primitives with a multiresolution hash encoding. ACM Transactions on Graphics (ToG)  \textbf{41}(4),  1--15 (2022)

\bibitem{pfister2000surfels}
Pfister, H., Zwicker, M., Van~Baar, J., Gross, M.: Surfels: Surface elements as rendering primitives. In: Proceedings of the 27th annual conference on Computer graphics and interactive techniques. pp. 335--342 (2000)

\bibitem{pittaluga2019revealing}
Pittaluga, F., Koppal, S.J., Kang, S.B., Sinha, S.N.: Revealing scenes by inverting structure from motion reconstructions. In: Proceedings of the IEEE/CVF Conference on Computer Vision and Pattern Recognition. pp. 145--154 (2019)

\bibitem{rakhimov2022npbg++}
Rakhimov, R., Ardelean, A.T., Lempitsky, V., Burnaev, E.: Npbg++: Accelerating neural point-based graphics. In: Proceedings of the IEEE/CVF Conference on Computer Vision and Pattern Recognition. pp. 15969--15979 (2022)

\bibitem{ramon2021h3d}
Ramon, E., Triginer, G., Escur, J., Pumarola, A., Garcia, J., Giro-i Nieto, X., Moreno-Noguer, F.: H3d-net: Few-shot high-fidelity 3d head reconstruction. In: Proceedings of the IEEE/CVF International Conference on Computer Vision. pp. 5620--5629 (2021)

\bibitem{roessle2022dense}
Roessle, B., Barron, J.T., Mildenhall, B., Srinivasan, P.P., Nie{\ss}ner, M.: Dense depth priors for neural radiance fields from sparse input views. In: Proceedings of the IEEE/CVF Conference on Computer Vision and Pattern Recognition. pp. 12892--12901 (2022)

\bibitem{ronneberger2015u}
Ronneberger, O., Fischer, P., Brox, T.: U-net: Convolutional networks for biomedical image segmentation. In: Medical Image Computing and Computer-Assisted Intervention--MICCAI 2015: 18th International Conference, Munich, Germany, October 5-9, 2015, Proceedings, Part III 18. pp. 234--241. Springer (2015)

\bibitem{ruckert2022adop}
R{\"u}ckert, D., Franke, L., Stamminger, M.: Adop: Approximate differentiable one-pixel point rendering. ACM Transactions on Graphics (ToG)  \textbf{41}(4),  1--14 (2022)

\bibitem{sabour2023robustnerf}
Sabour, S., Vora, S., Duckworth, D., Krasin, I., Fleet, D.J., Tagliasacchi, A.: Robustnerf: Ignoring distractors with robust losses. In: Proceedings of the IEEE/CVF Conference on Computer Vision and Pattern Recognition. pp. 20626--20636 (2023)

\bibitem{schonberger2016structure}
Schonberger, J.L., Frahm, J.M.: Structure-from-motion revisited. In: Proceedings of the IEEE conference on computer vision and pattern recognition. pp. 4104--4113 (2016)

\bibitem{simonyan2014very}
Simonyan, K., Zisserman, A.: Very deep convolutional networks for large-scale image recognition. arXiv preprint arXiv:1409.1556  (2014)

\bibitem{sun2022direct}
Sun, C., Sun, M., Chen, H.T.: Direct voxel grid optimization: Super-fast convergence for radiance fields reconstruction. In: Proceedings of the IEEE/CVF Conference on Computer Vision and Pattern Recognition. pp. 5459--5469 (2022)

\bibitem{suvorov2022resolution}
Suvorov, R., Logacheva, E., Mashikhin, A., Remizova, A., Ashukha, A., Silvestrov, A., Kong, N., Goka, H., Park, K., Lempitsky, V.: Resolution-robust large mask inpainting with fourier convolutions. In: Proceedings of the IEEE/CVF winter conference on applications of computer vision. pp. 2149--2159 (2022)

\bibitem{szegedy2015going}
Szegedy, C., Liu, W., Jia, Y., Sermanet, P., Reed, S., Anguelov, D., Erhan, D., Vanhoucke, V., Rabinovich, A.: Going deeper with convolutions. In: Proceedings of the IEEE conference on computer vision and pattern recognition. pp.~1--9 (2015)

\bibitem{tancik2022block}
Tancik, M., Casser, V., Yan, X., Pradhan, S., Mildenhall, B., Srinivasan, P.P., Barron, J.T., Kretzschmar, H.: Block-nerf: Scalable large scene neural view synthesis. In: Proceedings of the IEEE/CVF Conference on Computer Vision and Pattern Recognition. pp. 8248--8258 (2022)

\bibitem{tancik2023nerfstudio}
Tancik, M., Weber, E., Ng, E., Li, R., Yi, B., Wang, T., Kristoffersen, A., Austin, J., Salahi, K., Ahuja, A., et~al.: Nerfstudio: A modular framework for neural radiance field development. In: ACM SIGGRAPH 2023 Conference Proceedings. pp. 1--12 (2023)

\bibitem{tewari2020state}
Tewari, A., Fried, O., Thies, J., Sitzmann, V., Lombardi, S., Sunkavalli, K., Martin-Brualla, R., Simon, T., Saragih, J., Nie{\ss}ner, M., et~al.: State of the art on neural rendering. In: Computer Graphics Forum. vol.~39, pp. 701--727. Wiley Online Library (2020)

\bibitem{thies2019deferred}
Thies, J., Zollh{\"o}fer, M., Nie{\ss}ner, M.: Deferred neural rendering: Image synthesis using neural textures. Acm Transactions on Graphics (TOG)  \textbf{38}(4),  1--12 (2019)

\bibitem{turki2022mega}
Turki, H., Ramanan, D., Satyanarayanan, M.: Mega-nerf: Scalable construction of large-scale nerfs for virtual fly-throughs. In: Proceedings of the IEEE/CVF Conference on Computer Vision and Pattern Recognition. pp. 12922--12931 (2022)

\bibitem{vu2011high}
Vu, H.H., Labatut, P., Pons, J.P., Keriven, R.: High accuracy and visibility-consistent dense multiview stereo. IEEE transactions on pattern analysis and machine intelligence  \textbf{34}(5),  889--901 (2011)

\bibitem{waechter2014let}
Waechter, M., Moehrle, N., Goesele, M.: Let there be color! large-scale texturing of 3d reconstructions. In: Computer Vision--ECCV 2014: 13th European Conference, Zurich, Switzerland, September 6-12, 2014, Proceedings, Part V 13. pp. 836--850. Springer (2014)

\bibitem{wang2021patchmatchnet}
Wang, F., Galliani, S., Vogel, C., Speciale, P., Pollefeys, M.: Patchmatchnet: Learned multi-view patchmatch stereo. In: Proceedings of the IEEE/CVF conference on computer vision and pattern recognition. pp. 14194--14203 (2021)

\bibitem{wang2023f2}
Wang, P., Liu, Y., Chen, Z., Liu, L., Liu, Z., Komura, T., Theobalt, C., Wang, W.: F2-nerf: Fast neural radiance field training with free camera trajectories. In: Proceedings of the IEEE/CVF Conference on Computer Vision and Pattern Recognition. pp. 4150--4159 (2023)

\bibitem{wang2022mvster}
Wang, X., Zhu, Z., Huang, G., Qin, F., Ye, Y., He, Y., Chi, X., Wang, X.: Mvster: Epipolar transformer for efficient multi-view stereo. In: European Conference on Computer Vision. pp. 573--591. Springer (2022)

\bibitem{wei2021aa}
Wei, Z., Zhu, Q., Min, C., Chen, Y., Wang, G.: Aa-rmvsnet: Adaptive aggregation recurrent multi-view stereo network. In: Proceedings of the IEEE/CVF International Conference on Computer Vision. pp. 6187--6196 (2021)

\bibitem{wu2020multi}
Wu, M., Wang, Y., Hu, Q., Yu, J.: Multi-view neural human rendering. In: Proceedings of the IEEE/CVF Conference on Computer Vision and Pattern Recognition. pp. 1682--1691 (2020)

\bibitem{xie2023s3im}
Xie, Z., Yang, X., Yang, Y., Sun, Q., Jiang, Y., Wang, H., Cai, Y., Sun, M.: S3im: Stochastic structural similarity and its unreasonable effectiveness for neural fields. arXiv preprint arXiv:2308.07032  (2023)

\bibitem{xu2022point}
Xu, Q., Xu, Z., Philip, J., Bi, S., Shu, Z., Sunkavalli, K., Neumann, U.: Point-nerf: Point-based neural radiance fields. In: Proceedings of the IEEE/CVF Conference on Computer Vision and Pattern Recognition. pp. 5438--5448 (2022)

\bibitem{yao2018mvsnet}
Yao, Y., Luo, Z., Li, S., Fang, T., Quan, L.: Mvsnet: Depth inference for unstructured multi-view stereo. In: Proceedings of the European conference on computer vision (ECCV). pp. 767--783 (2018)

\bibitem{yeshwanth2023scannet++}
Yeshwanth, C., Liu, Y.C., Nie{\ss}ner, M., Dai, A.: Scannet++: A high-fidelity dataset of 3d indoor scenes. In: Proceedings of the IEEE/CVF International Conference on Computer Vision. pp. 12--22 (2023)

\bibitem{yu2019free}
Yu, J., Lin, Z., Yang, J., Shen, X., Lu, X., Huang, T.S.: Free-form image inpainting with gated convolution. In: Proceedings of the IEEE/CVF international conference on computer vision. pp. 4471--4480 (2019)

\bibitem{zamir2022restormer}
Zamir, S.W., Arora, A., Khan, S., Hayat, M., Khan, F.S., Yang, M.H.: Restormer: Efficient transformer for high-resolution image restoration. In: Proceedings of the IEEE/CVF conference on computer vision and pattern recognition. pp. 5728--5739 (2022)

\bibitem{zamir2021multi}
Zamir, S.W., Arora, A., Khan, S., Hayat, M., Khan, F.S., Yang, M.H., Shao, L.: Multi-stage progressive image restoration. In: Proceedings of the IEEE/CVF conference on computer vision and pattern recognition. pp. 14821--14831 (2021)

\bibitem{zhang2021gigamvs}
Zhang, J., Zhang, J., Mao, S., Ji, M., Wang, G., Chen, Z., Zhang, T., Yuan, X., Dai, Q., Fang, L.: Gigamvs: a benchmark for ultra-large-scale gigapixel-level 3d reconstruction. IEEE Transactions on Pattern Analysis and Machine Intelligence  \textbf{44}(11),  7534--7550 (2021)

\bibitem{zhang2020nerf++}
Zhang, K., Riegler, G., Snavely, N., Koltun, V.: Nerf++: Analyzing and improving neural radiance fields. arXiv preprint arXiv:2010.07492  (2020)

\bibitem{zhang2018unreasonable}
Zhang, R., Isola, P., Efros, A.A., Shechtman, E., Wang, O.: The unreasonable effectiveness of deep features as a perceptual metric. In: Proceedings of the IEEE conference on computer vision and pattern recognition. pp. 586--595 (2018)

\bibitem{zhang2024papr}
Zhang, Y., Peng, S., Moazeni, A., Li, K.: Papr: Proximity attention point rendering. Advances in Neural Information Processing Systems  \textbf{36} (2024)

\end{thebibliography}
\newpage

\appendix
\label{sec:appendix}

\section{Further Discussions}

\begin{figure}[t]
\centering
\includegraphics[width=0.8\linewidth]{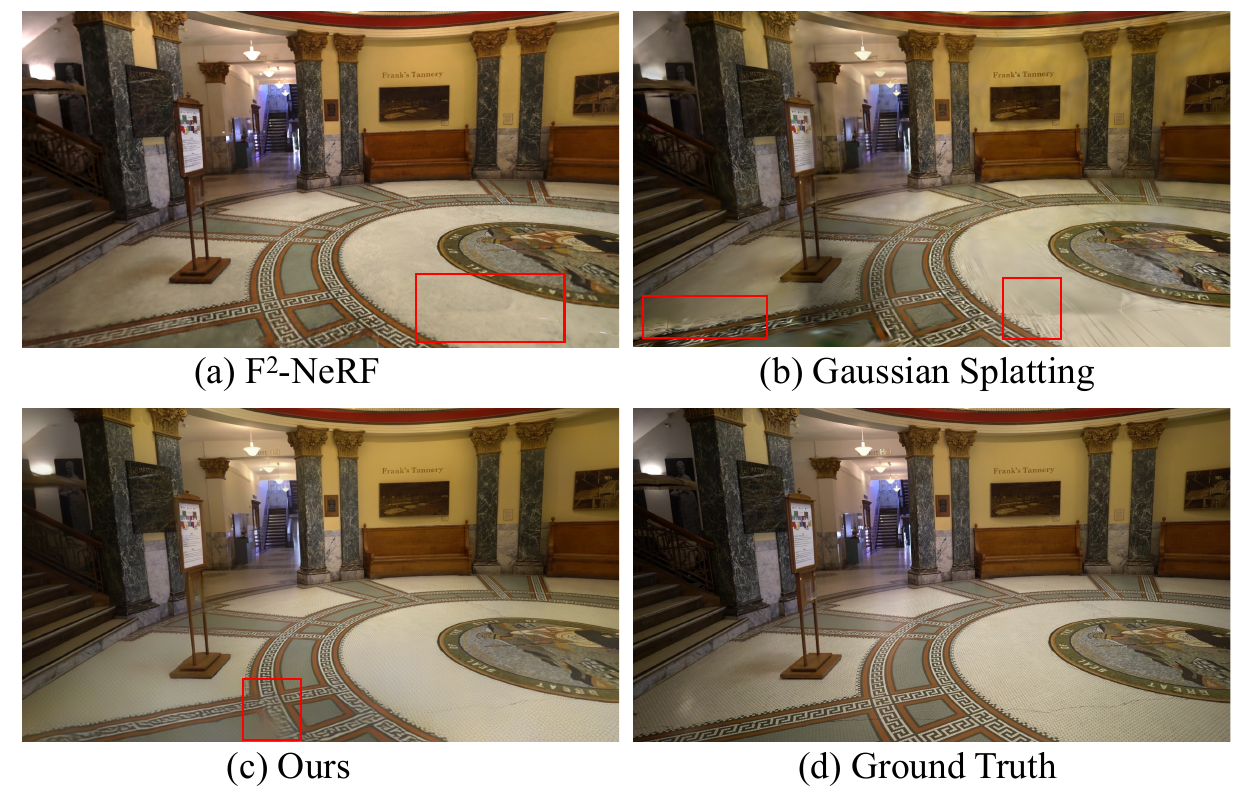}
\caption{Comparison on \textit{Museum} of T\&T dataset~\cite{knapitsch2017tanks} to showcase the typically different sources of noise due to the fundamental differences between different types of methods. Zoom in for best view.}
\label{fig:analysis}
\end{figure}

By extensive experiments, we have demonstrated the promising potential of point-based methods for NVS, especially the great generalizability and robustness of RPBG on varying scenes, with perceptually satisfactory rendering results.
We make a visual comparison on \textit{Museum} of T\&T dataset~\cite{knapitsch2017tanks} in \cref{fig:analysis} to demonstrate the varying sources of rendering artifacts and thus further discuss the fundamental differences between RF-based and point-based methods.

The privilege of adopting triangulated point clouds as the scene representation has been partially discussed in the main paper, as the points have contained all the verified co-visibility information across images.
Besides, the point-based representation enables certain editing of the target scene, which is more difficult for RF-based methods.
Please refer to \cref{sec:edit} for the cases of scene editing.

We also consider the convolutional patch-wise rendering scheme by RPBG plays an important role in achieving perceptually good renderings.
A similar ideology is explored in \cite{xie2023s3im} by enforcing structural supervision on a group of rendered pixels.

As the framed area in \cref{fig:analysis}(a) shows, the rendering noise is mainly caused by under-representation of RF, which is further due to the sparsity of input views (lack of ray intersections).
RF-based methods aim to represent the target scene loyally, where each inquiry is supposed to be a frank reflection of local optical properties.
In this way, RF-based methods render an image in pixels without considering the context information, establishing better pixel-to-pixel correspondence (thus higher PSNR).

We would like to in particular mention a series of RF-based methods, \eg, MVSNeRF~\cite{chen2021mvsnerf}, DS-NeRF~\cite{deng2022depth}, Point-NeRF~\cite{xu2022point}, DDP~\cite{roessle2022dense}, which incorporate geometric prior information for optimizing NeRFs.
They either adopt more explicit 3D proxies~\cite{chen2021mvsnerf,xu2022point} than RFs, or enforce supervision on the rendered depth~\cite{deng2022depth,roessle2022dense} to accelerate reconstruction or handle sparse views.
However, their rendering scheme is still RF-based volume rendering, leaving the relevant drawbacks remain.
For the readers' information, we also include the evaluation of DS-NeRF~\cite{deng2022depth} in \cref{tab:openmvs_dsnerf}.

\section{Dense Triangulation}

We here elaborate the details of the dense triangulation procedure to obtain the point clouds.

Recall that the NVS datasets consist of images and corresponding camera parameters (intrinsics and extrinsics).
To ensure the alignment between the poses and the reconstructed point clouds, we first triangulate sparse SIFT~\cite{lowe2004distinctive} points with COLMAP~\cite{schonberger2016structure}, where we only optimize the 3D coordinates, leaving the camera parameters frozen.

Based on the sparse triangulation, we follow the view selection strategy in \cite{yao2018mvsnet}, and choose 4 neighboring images with the best co-visibility for each image.
Then we estimate a depth map for each image, by aid of the top-4 neighboring images, with AA-RMVSNet~\cite{wei2021aa}.
The per-view depth maps are filtered and fused to obtain the final 3D point cloud.
We select AA-RMVSNet for its high memory-efficiency that allows a large batch size and RPBG is supposed to work fine with other off-the-self MVS methods.

For the scene of \textit{Building}, which consists of the most images among all the datasets (1940 images) we apply for quantitative experiments, the reconstruction can be done within one hour. 
With better engineering optimized algorithms, \eg, OpenMVS~\cite{openmvs2020}, the point cloud densification can be even faster.

Note that we leverage a point cloud augmentation strategy to relax the requirements of triangulated points.
More details will be covered in \cref{sec:edit}.

\section{Point Cloud}

In addition to the ablation study, we provide some further analysis and results relevant to the point-based proxy RPBG adopts, including the effectiveness of the point cloud augmentation strategy, the analysis of RPBG applied with random initialized points, RPBG's additional properties of automatic handling dynamic objects and scene editing.

\begin{figure}[t]
\centering
\includegraphics[width=0.95\linewidth]{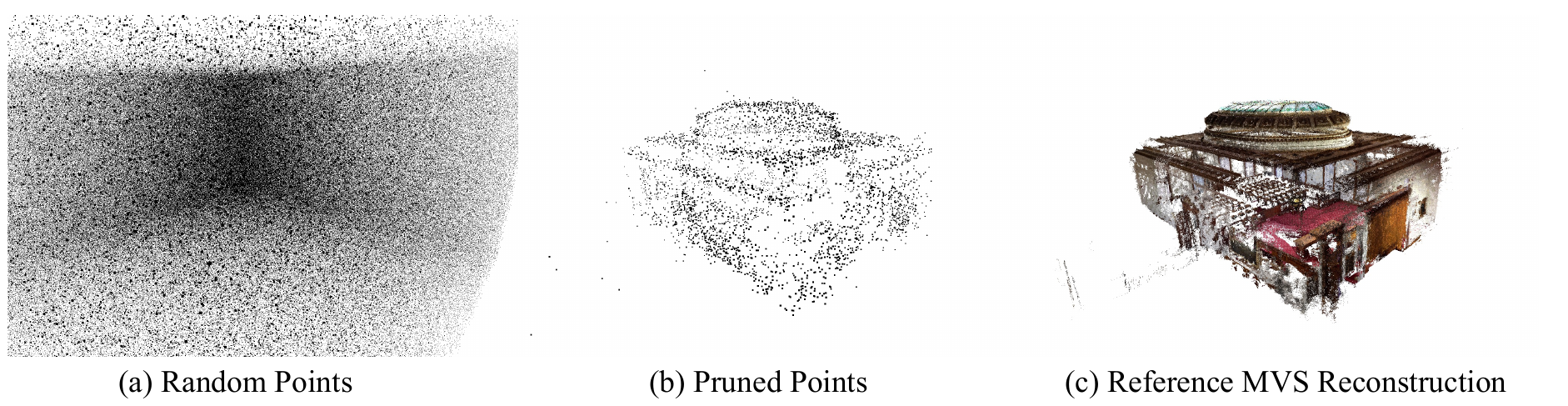}
\caption{The randomly initialized point cloud can be pruned to the coarse scene geometry. The example is \textit{Courtroom} from T\&T dataset~\cite{knapitsch2017tanks}.}
\label{fig:random}
\end{figure}

\paragraph{Point Cloud Augmentation}
The detailed augmentation steps are as \cref{alg:aug}.
\begin{algorithm}
\caption{Point Cloud Augmentation}\label{alg:aug}
\begin{algorithmic}[1]
\State \textbf{Input:} The point cloud $\{X\}$ to be augmented
\For{a given number of times}
\State Sample one existing point $X=(x,y,z)$ randomly from $\{X\}$
\State Form a 3D Gaussian distribution $G$ with its mean value $\mu=X$
\State Sample a new point $X'$ from $G$
\EndFor
\State Train RPBG with the point cloud $\{X\}\cup \{X'\}$
\For{each $X_i$ in $\{X'\}$}
    \State{Retrieve its neural texture $\mathbf{T}(X_i)$}
    \State Approximate its pseudo density $\sigma_i = \sum |\mathbf{T}(X_i)|$
    \If{$\sigma_i < \sigma_{\textrm{threshold}}$}
    \State{Discard $X_i$ from $\{X'\}$}
    \EndIf
\EndFor
\State \textbf{Output:} The augmented point cloud $\{X\}\cup \{X'\}$
\end{algorithmic}
\end{algorithm}
On the scene of \textit{Church}, we study the impact of such strategy applied on sparsely triangulated points multiple times.
As is shown in \cref{tab:aug_times}, the first round of sampling and pruning brings the largest performance gain, and a larger gain is observed when applying to the SfM-initialized triangulation, which is much sparser compared to the MVS-initialized one.
Note that the augmentation is optional and thus not performed on well triangulated scenes for the sake of time only.

\begin{table}[t]
    \centering
        \caption{Quantitative metrics when applying the point cloud augmentation strategy on both the sparsely and the densely triangulated points for different times.}
    \resizebox{0.6\linewidth}{!}{
    \begin{tabular}{c|ccc|ccc}
        \hline
        \multirow{2}{*}{\#\textit{Iters}} & \multicolumn{3}{c|}{\textit{\textbf{SfM Init.}}} & \multicolumn{3}{c}{\textit{\textbf{MVS Init.}}}\\
         & PSNR$\uparrow$ & SSIM$\uparrow$ & LPIPS$\downarrow$ & PSNR$\uparrow$ & SSIM$\uparrow$ & LPIPS$\downarrow$\\
        \hline
        0 & 21.41 & 0.750 & 0.318 & 23.16 &	0.809 & 0.243\\
        1 & 21.86 & 0.759 & 0.302 & 23.33 & 0.818 & 0.239\\
        2 & 21.85 & 0.761 & 0.303 & 23.36 &	0.814 & 0.241\\
        \hline
        $\Delta_{0\to 2}$ & +0.45 & +0.011 & -0.015 & +0.20 & +0.005 & -0.002\\
        \hline
    \end{tabular}
    }
    \label{tab:aug_times}
\end{table}

\paragraph{Random Point Cloud}
Since we have demonstrated in the main paper that RPBG is able to perform re-rendering even with a randomly initialized point cloud taken as input.
Empirically, we find that by applying the spatial pruning strategy to the random point cloud (by thresholding point-wise $\sigma > 180$ in this case), the point cloud shrinks to a shape similar to the actual geometry, as is illustrated in \cref{fig:random}(b).
It suggests that when neurally re-rendering, the network is able to implicitly verify the occupancy of each rasterized point and if a point is observed with poor multi-view consistency, it is more likely to be considered as an invalid point.
The attempt of pruning random points is considered as an extreme case explaining how the point cloud augmentation strategy of RPBG manages to alleviate the problem of patchy or erroneous triangulation.

\paragraph{Dynamic Objects}

\begin{figure}[t]
\centering
\includegraphics[width=0.9\linewidth]{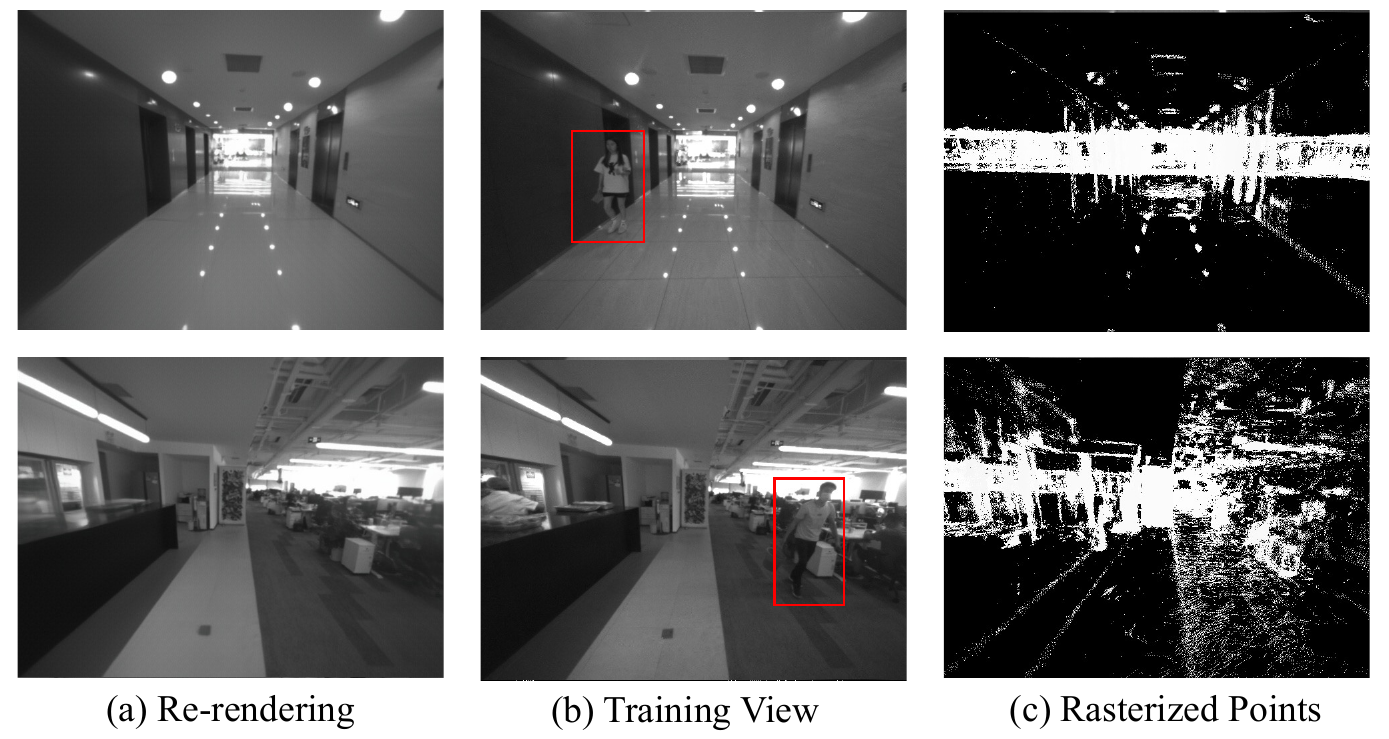}
\caption{Automatic removal of dynamic objects with RPBG on self-collected data. The points of the dynamic objects are not triangulated for they do not meet the static scene assumption.}
\label{fig:object}
\end{figure}

The RF-based methods are sensitive to dynamic objects and require either data pre-processing, \eg, masking by manual labeling and semantic segmentation, or modeling of such ambiguity or uncertainty~\cite{sabour2023robustnerf}, to aid the RF's optimization.
As for RPBG, the robustness against transient objects is trivially achieved since they are typically not reconstructed in SfM or MVS for not satisfying the static scene assumption.
By experiments, we discover that RPBG is robust to such dynamic objects and able to automatically such objects in the training views when re-rendering (\cref{fig:object}), which suggests that multi-view consistency is implicitly enforced during training and the renderer tends to restore the most consensual re-rendering.

\begin{figure}[t]
\centering
\includegraphics[width=0.8\linewidth]{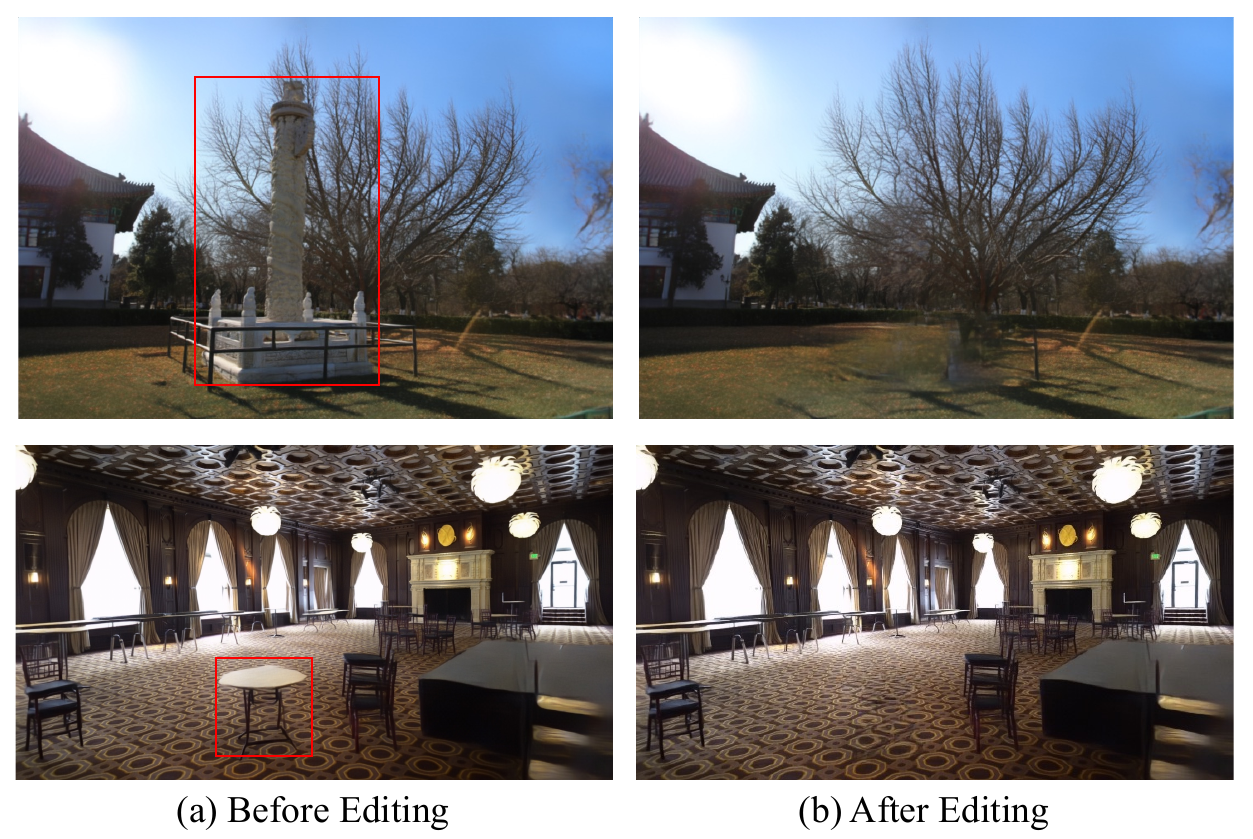}
\caption{Scene editing with RPBG on \textit{DayaTemple} of GigaMVS dataset~\cite{zhang2021gigamvs} and \textit{Ballroom} of T\&T dataset~\cite{knapitsch2017tanks}.}
\label{fig:edit}
\end{figure}

\paragraph{Scene Editing}\label{sec:edit}
As RPBG is a point-based pipeline, where an explicit 3D geometry is adopted for re-rendering, similar to previous point-based alternatives~\cite{aliev2020neural,zhang2024papr}, it allows certain scene editing and manipulation.
We give two examples in \cref{fig:edit}.
By removing the points, along with the point-bounded features, RPBG manages to re-render the edited scene, yet with some artifacts observed.
It is because in RPBG, we enhance the context exchange among rasterized points by DAC, where each point does not solely represent its local optical property.
Besides, it is also observed that FFC~\cite{fuoli2021fourier} may lead to repetitive artifacts at incomplete regions, as also can be found in the inpainted images by LaMa~\cite{suvorov2022resolution}.

\section{More Quantitative Results}
\label{sec:more_quantitative}

\paragraph{Traditional Reconstruction}
For a more comprehensive comparison, we also include OpenMVS~\cite{openmvs2020}, as a traditional pipeline~\cite{barnes2009patchmatch,jancosek2014exploiting,vu2011high,waechter2014let} that reconstructs textured mesh models to get rendered at arbitrary novel views.
We compare the results on \textit{Auditorium}, \textit{Ballroom}, and \textit{Courtroom} (\cref{tab:openmvs_dsnerf}) as they are inside-out scenes to avoid the negative impact of background.

\begin{table}[t]
    \centering
    \caption{Additional quantitative results on \textit{Auditorium}, \textit{Ballroom}, and \textit{Courtroom} of T\&T dataset~\cite{knapitsch2017tanks}. 
    The scores of F$^2$-NeRF~\cite{wang2023f2}, NPBG~\cite{aliev2020neural}, Gaussian Splatting~\cite{kerbl20233d} and RPBG are provided for reference.
    PSNR$\uparrow$/SSIM$\uparrow$/LPIPS$\downarrow$}
    \resizebox{0.8\linewidth}{!}{
    \begin{tabular}{l|c|c|c}
        \hline
        \textbf{Method} & \textit{\textbf{Auditorium}} & \textit{\textbf{Ballroom}} & \textit{\textbf{Courtroom}} \\
        \hline
        OpenMVS~\cite{openmvs2020} & 16.81/0.688/0.404 &	14.69/0.336/0.486 & 14.92/0.472/0.430\\
        DS-NeRF~\cite{deng2022depth} & 16.29/0.542/0.612 & 14.74/0.668/0.570 & 14.62/0.491/0.616 \\
        F$^2$-NeRF~\cite{wang2023f2} & 20.36/0.843/0.329 &	22.21/0.706/0.328 & 20.13/0.672/0.425 \\
        NPBG~\cite{aliev2020neural} & 22.05/0.814/0.375 &	21.04/0.681/0.330 & 20.99/0.681/0.386 \\
        {\footnotesize Gaussian Splatting~\cite{kerbl20233d}} & 23.82/0.868/0.288 & 22.96/0.769/0.227 & 22.43/0.765/0.278 \\
        \hline
        \textbf{Ours} & 25.08/0.888/0.245 & 23.36/0.782/0.217 & 23.22/0.781/0.249	\\

        \hline
    \end{tabular}
    }
    \label{tab:openmvs_dsnerf}
\end{table}

\paragraph{Geometry-bounded NeRF}

In RPBG, the scene parameterization relies on the sparse/dense triangulation which incorporates estimated depth maps by SfM/MVS.
To analogize this parameterization from the perspective of NeRF, we also evaluate DS-NeRF~\cite{deng2022depth} on the aforementioned inside-out scenes in \cref{tab:openmvs_dsnerf}.

\begin{figure}[t]
\centering
\includegraphics[width=\linewidth]{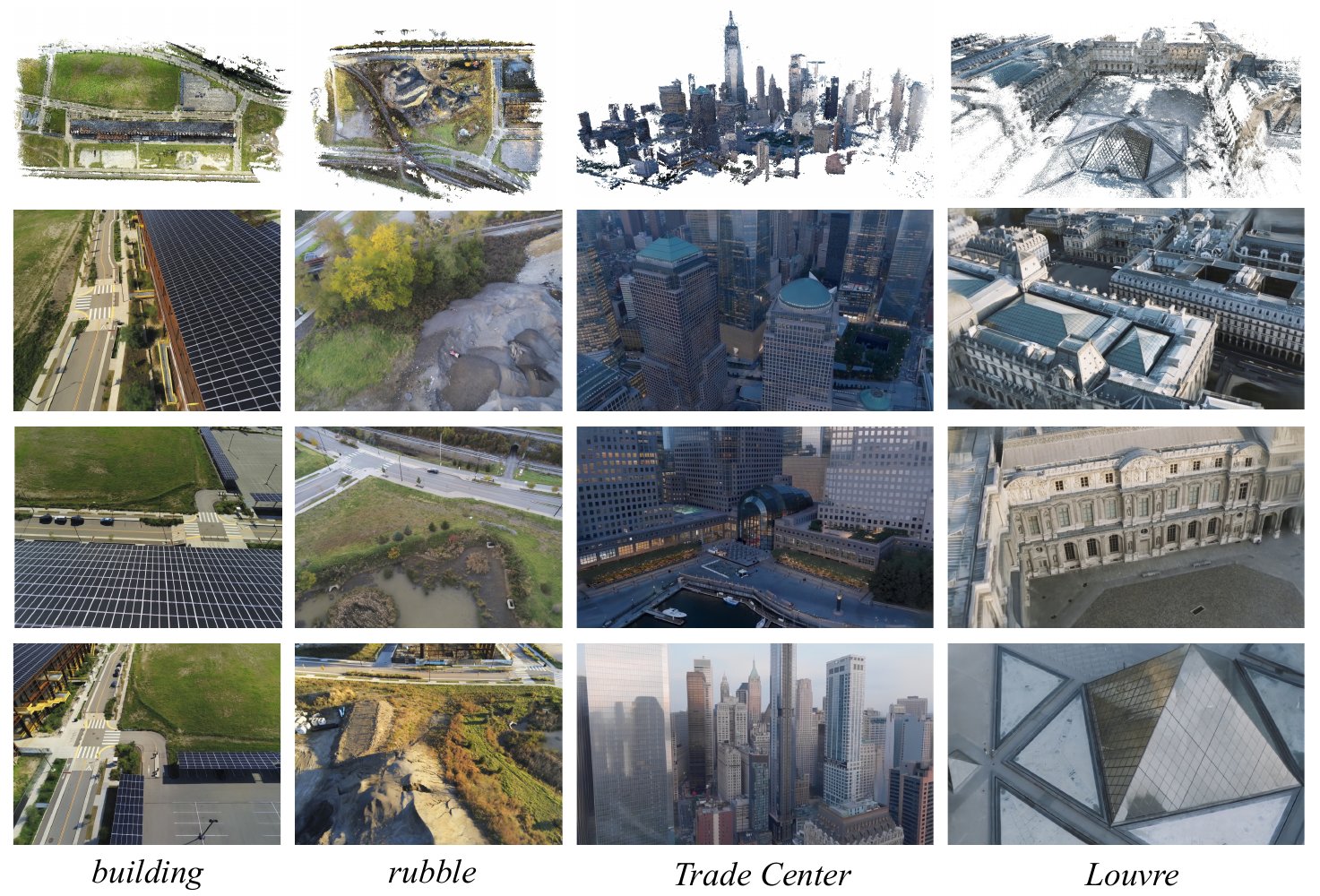}
\caption{Results of RPBG on aerial scenes, \ie, Mill19 dataset~\cite{turki2022mega} and OMMO dataset~\cite{lu2023large}.}
\label{fig:mill19_ommo}
\end{figure}

\paragraph{Densely Captured Dataset}
Though RPBG targets more generic scenes with casual settings, for the readers' information, we also evaluate RPBG with a densely captured dataset, NeRF-360 dataset~\cite{barron2022mip}, which is considered as ideal for training NVS, in \cref{fig:nerf360}.
Note that mip-NeRF-360~\cite{barron2022mip} is particularly designed for such cases and takes about $6\times$ time for training.

\begin{table}[t]
    \centering
    \caption{Additional quantitative results on NeRF-360 dataset~\cite{barron2022mip}. The provided methods~\cite{barron2021mip,zhang2020nerf++,barron2022mip} are typical unbounded NeRF variants.}
    \resizebox{0.8\linewidth}{!}{
    \begin{tabular}{l|c|ccc|ccc}
        \hline
        \multirow{2}{*}{\textbf{Method}} & GPU & \multicolumn{3}{c|}{\textit{\textbf{NeRF-360 outdoor}}} & \multicolumn{3}{c}{\textit{\textbf{NeRF-360 indoor}}}\\
        & Hours &  PSNR$\uparrow$ & SSIM$\uparrow$ & LPIPS$\downarrow$ & PSNR$\uparrow$ & SSIM$\uparrow$ & LPIPS$\downarrow$\\
        \hline
        mip-NeRF~\cite{barron2021mip} & 22 & 22.65 &	0.505 &	0.484 & 26.98 & 0.798 &	0.360 \\
        NeRF++~\cite{zhang2020nerf++} & 66 & 23.77 &	0.585 &	0.401 & 28.05 & 0.836 &	0.309\\
        mip-NeRF-360~\cite{barron2022mip} & 48 & \best{25.92} &	\best{0.747} &	\best{0.244} & \best{31.72} & \best{0.917} & 
        \secondbest{0.180}\\
        \textbf{Ours} & 8 & \secondbest{24.72} &	\secondbest{0.709} &	\secondbest{0.252} & \secondbest{28.76} &	\secondbest{0.898} & \best{0.140} \\
        \hline
    \end{tabular}
    }
    \label{fig:nerf360}
\end{table}

\paragraph{ScanNet++ Benchmark}
We also evaluate RPBG on the public benchmark of ScanNet++~\cite{yeshwanth2023scannet++} (Novel View Synthesis on DSLR Images).
ScanNet++ contains a wide variety of indoor scenes that are challenging for novel view synthesis for glossy and reflective materials and unseen poses captured independently of the training trajectory.
The results are shown in \cref{tab:scannetpp}. Note that the scores are all retrieved from the leaderboard.
RPBG outperforms all the baselines listed by the benchmark, with a particular good perceptual quality (LPIPS).

\begin{table}[t]
    \centering
    \caption{Benchmarking results on ScanNet++~\cite{yeshwanth2023scannet++}. The scores are retrieved by the evaluation system of the public benchmark.}
    \resizebox{0.5\linewidth}{!}{
    \begin{tabular}{l|c|c|c}
        \hline
        \textbf{Method} & PSNR$\uparrow$ & SSIM$\uparrow$ & LPIPS$\downarrow$ \\
        \hline
        Nerfacto~\cite{tancik2023nerfstudio} & \secondbest{24.05}	& 0.861	& 0.342\\
        Instant-NGP~\cite{muller2022instant} & 23.81 &	0.859 &	0.375\\
        Gaussian Splatting~\cite{kerbl20233d} & 	23.89	& \secondbest{0.871}	& \secondbest{0.319} \\
        \hline
        \textbf{Ours} & \best{24.36} &	\best{0.873} &	\best{0.280}\\

        \hline
    \end{tabular}
    }
    \label{tab:scannetpp}
\end{table}

\section{More Qualitative Results}

\paragraph{Mill19 and OMMO Results}
For the scenes in Mill19~\cite{turki2022mega} and OMMO~\cite{lu2023large}, we provide the triangulated points and visualized re-renderings in \cref{fig:mill19_ommo}.
Since RPBG represents the scene appearance with point-bounded features, it relieves users from partitioning large-scale data into smaller chunks, revealing the great scalability.
Besides, the DAC module is well suited to capture periodic structures, which are common in human-made environments~\cite{suvorov2022resolution}.

\begin{figure}[t]
\centering
\includegraphics[width=\linewidth]{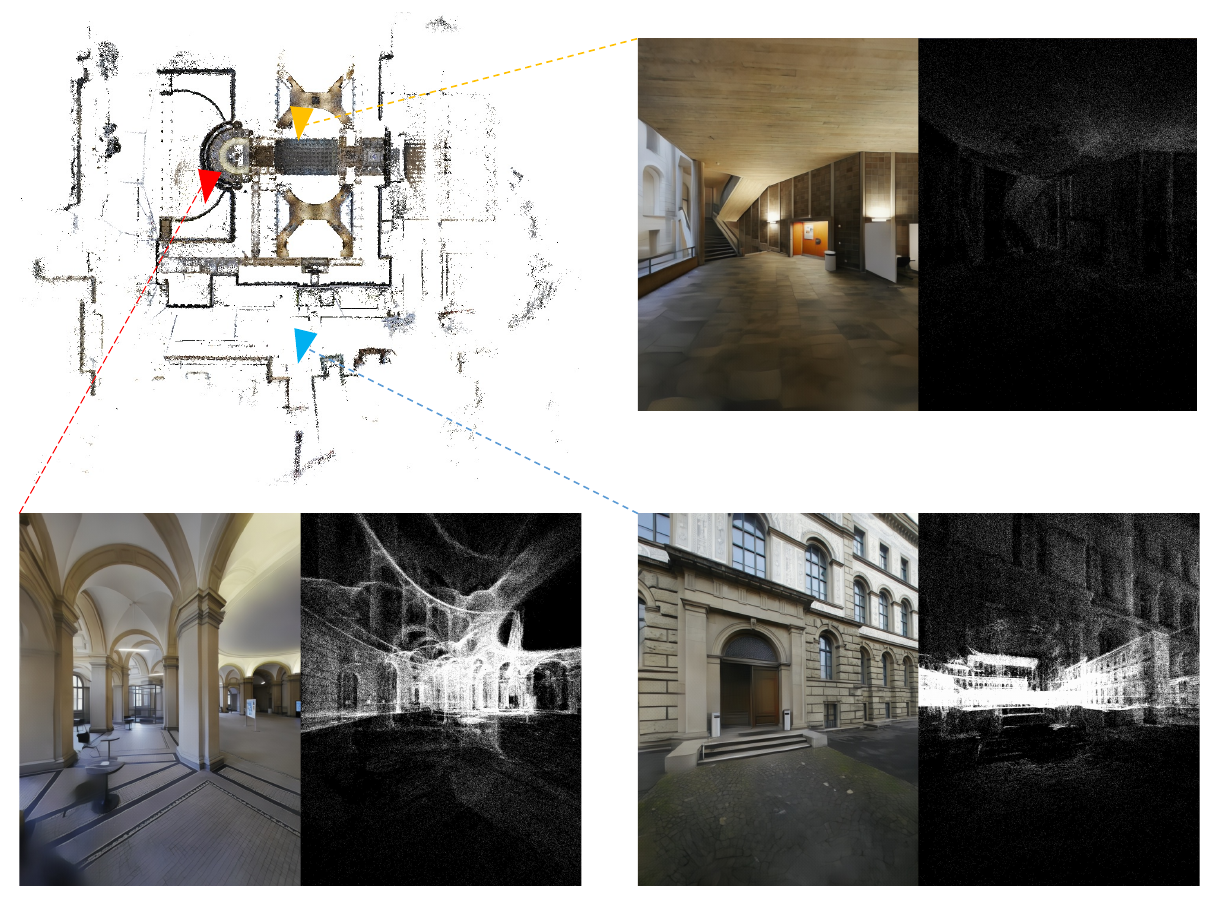}
\caption{Results of RPBG on ETH-MS dataset~\cite{eth_ms_visloc_2021}. The location and orientation of the sampled cameras are marked with different colors in the densely triangulated point cloud respectively.}
\label{fig:eth}
\end{figure}

\paragraph{ETH-MS Results}
We also test RPBG's capability of handling super-large-scale scenes on ETH-MS dataset~\cite{eth_ms_visloc_2021}, which is for visual localization in AR applications.
Its mapping set is captured by the 6-camera rig of a NavVis M6 mobile scanner, and contains 4914 images captured at the HG building of the campus of ETH Zurich, both in the main halls and on the sidewalk. 
The dataset is extremely challenging for NVS as its observations are very sparse and it exhibits many self-similarities and symmetric structures.
The triangulated dense point cloud as well as three novel views absent in the training set is demonstrated in \cref{fig:eth}.
Note that, RPBG also adopts the exactly identical settings without any partition of data.
RPBG achieves visually pleasing results even when the scene is extremely complicated, indicating that our re-rendering is robust to point sparsity and occlusion.

\section{Use of Existing Assets}

We here list all the existing assets used in this manuscript and would like to sincerely appreciate the maintainers of these open-source projects:

\begin{itemize}[noitemsep,topsep=0pt]
    \item NeRF~\cite{mildenhall2020nerf}, NeRF++~\cite{zhang2020nerf++}, and TensoRF~\cite{chen2022tensorf}: \url{https://github.com/ashawkey/torch-ngp}
    \item Mega-NeRF and Mill19 Dataset~\cite{turki2022mega}: \url{https://github.com/cmusatyalab/mega-nerf}
    \item F$^2$-NeRF and Free Dataset~\cite{wang2023f2}: \url{https://github.com/Totoro97/f2-nerf}
    \item NPBG~\cite{aliev2020neural} and NPBG++~\cite{rakhimov2022npbg++}: \url{https://github.com/rakhimovv/npbgpp}
    \item Gaussian Splatting~\cite{kerbl20233d}: \url{https://github.com/graphdeco-inria/gaussian-splatting}
    \item COLMAP~\cite{schonberger2016structure}: \url{https://colmap.github.io}
    \item OpenMVS~\cite{openmvs2020}: \url{https://github.com/cdcseacave/openMVS}
    \item AA-RMVSNet~\cite{wei2021aa}: \url{https://github.com/QT-Zhu/AA-RMVSNet}
    \item Tanks and Temples Benchmark~\cite{knapitsch2017tanks}: \url{https://www.tanksandtemples.org}
    \item OMMO Dataset~\cite{lu2023large}: \url{https://ommo.luchongshan.com}
    \item GigaMVS Benchmark~\cite{zhang2021gigamvs}: \url{https://www.gigavision.cn}
    \item ScanNet++ Benchmark~\cite{yeshwanth2023scannet++}: \url{https://kaldir.vc.in.tum.de/scannetpp/benchmark/nvs}
    \item ETH-MS Dataset~\cite{eth_ms_visloc_2021}: \url{https://github.com/cvg/visloc-iccv2021}
\end{itemize}

\end{document}